\documentclass[journal=jcisd8,manuscript=article,layoutreview]{achemso}
\usepackage[table,xcdraw]{xcolor}
\usepackage{amsmath}
\usepackage{dingbat}

\usepackage{hyperref}

\usepackage{tabularray}
 \usepackage{multirow}
 \usepackage{multicol}
\usepackage{tcolorbox}
\usepackage[T1]{fontenc}

\usepackage{titlesec}
\title{Sections and Chapters}

\usepackage[version=3]{mhchem}

\author{Afnan Sultan}
\affiliation[Saarland University]
{Data Driven Drug Design, Center for Bioinformatics, Saarland University, Germany}
\altaffiliation{Equal Contribution}

\author{Jochen Sieg}
\altaffiliation{Equal Contribution}

\author{Miriam Mathea}
\email{miriam.mathea@basf.com}
\affiliation[BASF]
{BASF SE, Ludwigshafen. Germany}

\author{Andrea Volkamer}
\email{volkamer@cs.uni-saarland.de}
\affiliation[Saarland University]
{Data Driven Drug Design, Center for Bioinformatics, Saarland University, Germany}

\title[An \textsf{achemso} demo]
  {Transformers for molecular property prediction: Lessons learned from the past five years}

\begin{document}

\begin{abstract}

Molecular Property Prediction (MPP) is vital for drug discovery, crop protection, and environmental science. Over the last decades, diverse computational techniques have been developed, from using simple physical and chemical properties and molecular fingerprints in statistical models and classical machine learning to advanced deep learning approaches.
In this review, we aim to distill insights from current research on employing transformer models for MPP. We analyze the currently available models and explore key questions that arise when training and fine-tuning a transformer model for MPP. These questions encompass the choice and scale of the pre-training data, optimal architecture selections, and promising pre-training objectives. Our analysis highlights areas not yet covered in current research, inviting further exploration to enhance the field's understanding. Additionally, we address the challenges in comparing different models, emphasizing the need for standardized data splitting and robust statistical analysis.       
\end{abstract}

\tableofcontents

\section{Introduction}

% Machine learning
Modern deep learning architectures are enabling applications across various domains. Architectures originally developed for computer vision \cite{krizhevsky2012imagenet, he2016deep} and language processing \cite{vaswani2017attention} are reshaping central life science fields \cite{jumper2021highly, baek2021accurate}. These successes raise high expectations for the long-standing challenge of small molecule property prediction. Molecular property prediction (MPP) methods have a wide range of applications, including drug discovery, material science, crop protection, and environmental science. 
Accurate predictions of properties like on-target activity, physico-chemical properties, or (eco-)toxicity of the molecule \cite{wu2018moleculenet} can help researchers prioritize compounds for further experimentation, assess the potential toxicity of chemicals, optimize drug candidates, and design new materials with desired properties \cite{wieder2020compact, bender2022evaluation, deng2023systematic}.

% representations
A key problem in the field of molecular property prediction is to find a molecular representation that corresponds well with relevant properties \cite{deng2023systematic}. An interesting difference between deep learning methods to more traditional quantitative structure-activity relationship (QSAR) modeling \cite{tropsha2010best} is that representations can be learned from data instead of using fixed or human-engineered representations \cite{yang2019analyzing, deng2023systematic}. For decades, fixed representations, such as molecular descriptors and fingerprints, have been employed to describe a molecule by converting it into a fixed vector length for using it in machine learning algorithms. Molecular descriptors can include physico-chemical properties, topological indices, and quantum chemistry properties. They provide a straightforward representation but are limited due to a priori selection. 
A special type of descriptors are molecular fingerprints, usually binary bit strings representing the presence or absence of substructures or functional groups in a molecule\cite{shen2019molecular, david2020molecular}. Popular examples are the extended-connectivity fingerprint (ECFP) \cite{rogers2010extended}, which represents atom environments, and MACCS structural keys, which encode substructures \cite{durant2002reoptimization}.

% deep learning
Multiple deep learning architectures have been proposed for supervised learning for MPP \cite{walters2020applications, li2022deep}. Graph neural networks and graph convolutional neural networks \cite{wu2020comprehensive} can learn from molecular graphs representing atoms and their connections \cite{wieder2020compact, corso2024graph}. 
Another approach is the use of recurrent neural networks \cite{hochreiter1997long} to model sequential molecule data, such as SMILES \cite{weininger1988smiles} strings \cite{winter2019learning, deng2023systematic}. 
Generative models, like variational autoencoders \cite{kingma2019introduction}, generative adversarial networks \cite{goodfellow2014generative} and normalizing flow models \cite{rezende2015variational}, are also employed for molecular property prediction. These models learn to embed the input molecule data into a latent space to generate novel molecules with desired properties by sampling from the latent space representation \cite{winter2019learning, meyers2021novo, cheng2021molecular, deng2022artificial, bilodeau2022generative, anstine2023generative, pang2023deep}.
Transfer learning \cite{tan2018survey} is another promising technique. By training a deep learning model on one task with a sufficiently large data set of molecular structures and property labels, the model can learn general features transferable to a specific task with few labeled training data \cite{cai2020transfer, li2020inductive, dou2023machine}.

% SSL
However, labeled molecular property data sets are notoriously small, containing only hundreds or thousands of molecules \cite{wu2018moleculenet, yang2019analyzing, stanley2021fs, deng2023systematic}. This small size limits supervised learning algorithms to derive potent representations for all properties of interest. Self-supervised learning (SSL) \cite{liu2021self} has emerged as a potential solution to overcome this bottleneck by exploiting the large corpus of unlabeled molecular data, which has been explored on SMILES \cite{wang2019smiles, chithrananda2020chemberta, fabian2020molecular, irwin2022chemformer, ross2022large} and molecular graphs \cite{rong2020self, xie2022self, zang2023hierarchical}. SSL aims to learn general-purpose features or task-agnostic representations, meaning these representations can be reused in downstream tasks as proposed by BERT \cite{devlin2018bert}. SSL approaches the feature extraction problem with a pre-training task (or pretext task), for example, with the objective of predicting missing parts in a masked input data point \cite{ericsson2022self}. This approach is currently explored with great effort in the chemical domain \cite{wang2019smiles, chithrananda2020chemberta, fabian2020molecular, irwin2022chemformer, ross2022large, rong2020self, zang2023hierarchical, schwaller2019molecular}.
Still, standard machine learning models and features, like tree-based methods with fixed-size molecular fingerprints, remain strong baselines for real-world data sets \cite{yang2019analyzing, kimber2021deep, deng2023systematic}. This illustrates certain challenges that need to be overcome to exploit the full potential of deep learning models in this field.  

% Transformers
Transformers\cite{vaswani2017attention} are interesting architectures for molecular property prediction applications. They were developed originally for language processing and rely on sequential input and output. They have been a central component in the breakthrough of AlphaFold2 and other methods in the related field of protein structure prediction from the primary sequence \cite{jumper2021highly, baek2021accurate, lin2023evolutionary}. Proteins' sequential nature, i.e. their amino acid sequence, makes them an appropriate basis for developing protein language models.
Accordingly, the expectations for extending this success to property prediction of other biomolecules are raised. For small molecules, the widely-used SMILES \cite{weininger1988smiles} strings are an intuitive sequential molecular encoding ready to be used with language models. The attention mechanism \cite{bahdanau2014neural} used by transformers aims to learn context-dependent relationships and might be suited to capture complex non-additivities\cite{gogishvili2021nonadditivity} in molecular data. In addition, their encoder-decoder architecture is suitable for both molecular property prediction and molecule generation and they are the foundation for powerful architectures in computation linguistics \cite{devlin2018bert, brown2020language}. Transformers have also been shown to scale well to enormously large unlabeled data sets \cite{devlin2018bert, radford2019language}.   

In this review, we aim to give an overview of the current state of transformer models for molecular property prediction (MPP). We first present the transformer architecture, its variants from the language domain, and their adaptions for MPP, which we categorize by their architectural differences. In the second section, we overview the data sets used for training and evaluation and examine the reported prediction performances. We then review the choices to consider when implementing a transformer model for MPP and discuss architectural choices, training objectives, and data set sizes. Finally, we describe the challenges encountered during this review, highlighting the need for a unified benchmark and reporting, and we provide potential next steps for using transformers for MPP.  

\begin{tcolorbox}[width=\textwidth,colback={white},title={Take home messages for molecular transformers for property prediction tasks},colbacktitle=blue!50!white,coltitle=black]
\begin{tabular}{p{0.15\textwidth} p{0.8\textwidth}}
Performance & The transformer models from the language domain using mostly SMILES language show comparable performance to existing machine and deep learning models for MPP. \\
Scaling & The field needs a systematic analysis of the scaling of the number of model parameters and the pre-training data set size to avoid over- or under-training the model. \\
Pre-training & Data selection methods for pre-training can help with the model's generalizability while reducing the data set size. \\
Language & Preliminary results suggest that input representation is not a major factor, as different models have effectively used SMILES, SELFIES, circular fingerprints, and a simple list of atoms.\\
Domain knowledge & Current attempts to adopt domain-relevant pre-training objectives have shown promising effects. However, no analysis has been carried out yet for tokenization and positional encoding (PE). We believe that they need further investigations as they can help the process of model learning. Domain-relevant tokenization and 2D- or 3D-aware PE can provide better performance and/or explainability.\\
Benchmarks & Missing standardization for data splitting, statistical analysis, and reporting limit proper comparison of current literature methods. \\
\end{tabular}
\end{tcolorbox}

\section{The transformer model}

\begin{figure}
  \includegraphics[width=1.0\linewidth]{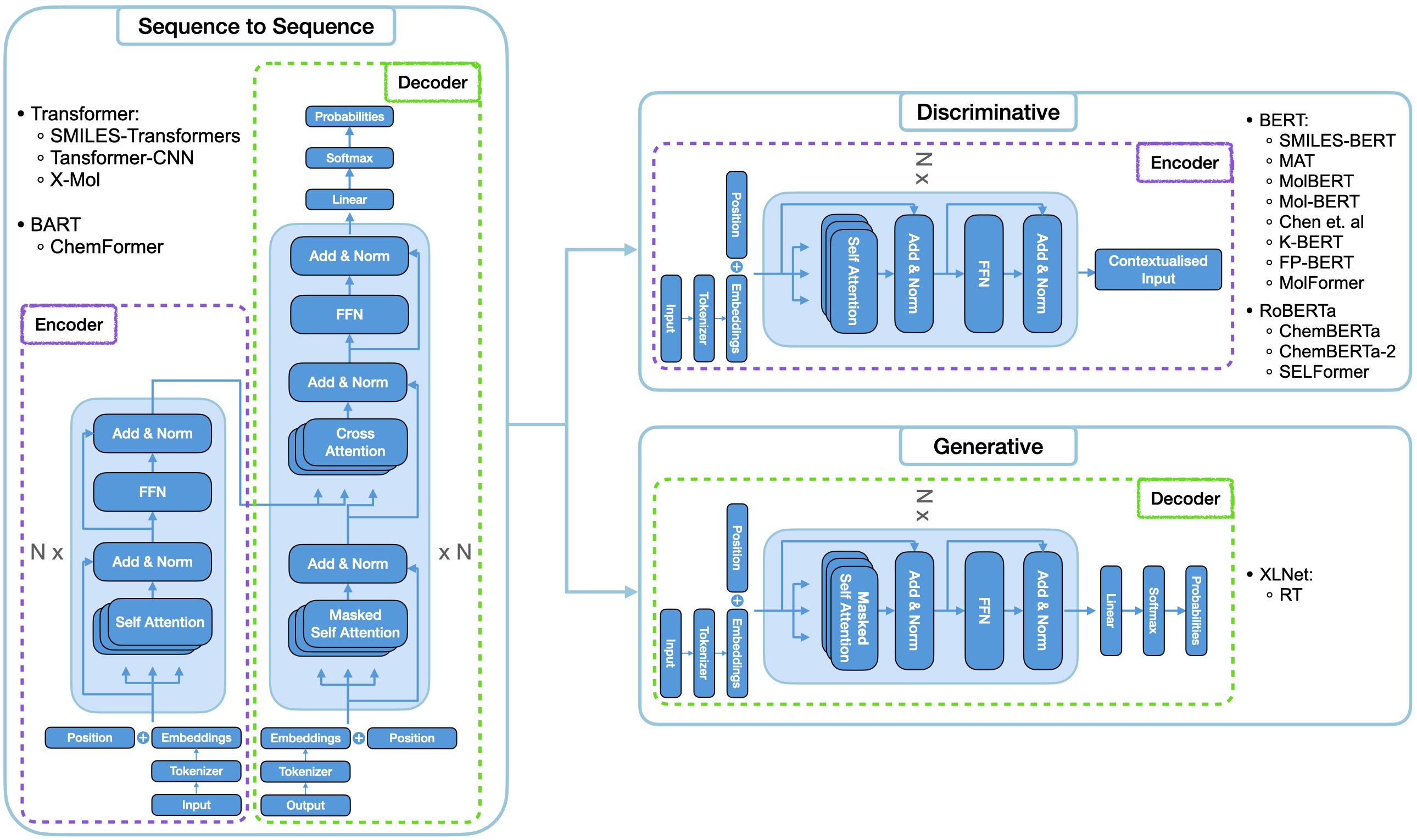}
  \caption{An overview of the transformer model and where each chemical language model fits. The transformer model with the encoder-decoder modules is a sequence-to-sequence model that can be used for tasks like reaction prediction. However, each module can be used independently to provide more specialized performance. For example, the encoder module can be used to predict properties, while the decoder module can be used to generate novel molecules.}
  \label{fig:transformers}
\end{figure}

The transformer model is a sequence-to-sequence model with the main aim of translating an input sequence into an output sequence. This concept has been utilized for different tasks in NLP, such as machine translation or text summarization \cite{vaswani2017attention}, as well as tasks in cheminformatics such as reaction prediction \cite{schwaller2019molecular} and molecular optimization \cite{he2022transformer}. In this section, we will explain the transformer model and some of its variants, followed by outlaying the chemical language models derived from it (Figure \ref{fig:transformers}).

\subsection{How transformer models work}
The transformer model proposed by Vaswani et al. \cite{vaswani2017attention} took the field of natural language processing (NLP) by storm as the current citation for this work is over 100K \cite{googleScholar}. The model consists of stacks of encoders and decoders that rely mainly on the previously used attention mechanism \cite{bahdanau2014neural}. This dispensed the need for using recurrent or convolutional neural networks (RNN and CNN), which were the SOTA prior to the transform model \cite{vaswani2017attention}. Relying mainly on the attention mechanism allowed for parallel processing of the input sequence, which was previously hindered, for example, due to the sequential nature of RNNs. This enabled a significant reduction in computation time as well as processing longer sequences which enables learning longer-range dependencies. 

The attention mechanism proposed by Vaswani et al. \cite{vaswani2017attention} relies on a scaled dot product of three linear transformations that are named the query (Q), key (K), and value (V) vectors. These vectors are of length $d_k$ which are calculated for each input sequence and attention is performed as follows:

\[Attention(Q, K,\ and\ V) = softmax(\dfrac{QK^T}{\sqrt{d_k}})V\]

Besides the attention mechanism, the transformer model comprises layer normalization steps that stabilize calculations and a feed-forward neural network which is the only source of non-linearity in the model.
The input to the encoder model is a tokenized sequence.  I.e., the input sequence is broken into composing units called tokens, which can be characters, words, subwords, etc., depending on the tokenization method (see Figure \ref{fig:transformers}).

The transformer model's encoder unit focuses on identifying a token's relationship to its surrounding tokens, including previous and following tokens, which is done via self-attention. 
However, the decoder subunit is a generative module that focuses on predicting the next token from strictly processing the previous tokens only.
The decoder uses masked self-attention for this task as the future words for each token are masked. Additionally, the decoder's generation task is offset by one token to allow for the output embeddings of the encoder module to be processed by the decoder module. At this step, a third attention type called cross-attention is performed, which computes the relationship of the decoder's generated tokens to the encoder's contextualized embeddings of the input tokens. This cross-attention step allows for both the encoder and the decoder to align their predictions between the input sequence and the desired output sequence. For example, in the case of the original transformer model \cite{vaswani2017attention}, the goal was machine translation between different languages.

In addition to reducing computation cost and detecting longer-range dependencies, the transformer model paved the way for the \textit{pre-training and fine-tuning paradigm} as proposed in BERT \cite{devlin2018bert}. In this paradigm, the model is trained on a large unlabeled data set with objectives that allow the model to form a general comprehension of the training domain. Afterward, the pre-trained model can be tuned on the labeled - but significantly smaller - downstream data set. The transformer models have enabled the training of foundational models that can easily be fine-tuned for numerous downstream data sets.

Although the architecture of the transformer model makes it a sequence-to-sequence model, the individual encoder and decoder subunits of the model have been exploited independently to perform discriminative or generative tasks (Figure \ref{fig:transformers}). For example, the BERT \cite{devlin2018bert} model is an encoder-only model and the famous GPT models are decoder-only models \cite{brown2020language, radford2019language}. Additionally, to improve the pre-training and fine-tuning paradigm, more models have proposed additional tweaks and objectives to the initial transformer model that showed improved performance or generalizability for specific tasks. 

The reviewed molecular transformer model articles have used some of these variants as base models. For example, Table \ref{tab:articles} shows that some models adopted the original transformer \cite{vaswani2017attention} model while others adopted the BERT \cite{devlin2018bert} model. 

\subsection{Adopted variants of the transformer model for MPP}

Although the transformer \cite{vaswani2017attention} model provided unprecedented performance in the NLP field, the following models proposed tweaks and objectives to help with a specific task or to improve the technique further. Some models became a defacto for their corresponding task, such as BERT \cite{devlin2018bert} for discriminative tasks. In the following, we provide a brief overview of the BERT \cite{devlin2018bert}, RoBERTa \cite{liu2019roberta}, BART \cite{lewis2020bart}, and XLNet \cite{yang2019xlnet} models that served as basis for the reviewed models.

\textbf{BERT} \cite{devlin2018bert} (Bidirectional Encoder Representations from Transformers) is an encoder model that was trained for Natural Language Understanding (NLU). It proposed two pre-training objectives that allow for better comprehension of the language: Masked language modeling (MLM) and Next Sentence Prediction (NSP). In MLM, a random percentage of the tokens in each input is replaced with a MASK token, and the model has to predict the correct embedding for these masked tokens. For NSP, the model is given a text pair (i.e., two segments of the document) and must classify whether they are consecutive in the original text or not. 

\textbf{RoBERTa} \cite{liu2019roberta} (Robustly optimized BERT Approach) is a replication of BERT with optimized hyperparameters. The key differences between RoBERTa and BERT lie in the training process. RoBERTa is trained with larger batch sizes, more data, and longer training duration. It also removes the next sentence prediction task used in BERT, as it was found to add minimal to no gain in performance. RoBERTa also uses dynamic masking, changing the masking pattern during pre-training, which should improve generalization.

\textbf{BART} \cite{lewis2020bart} (Bidirectional and AutoRegressive Transformers) is an encoder-decoder model that is meant as a denoising autoencoder. It is trained by corrupting the input text (i.e., adding noise to the input) and then reconstructing the original text. The model learns to generate high-quality output by learning to denoise the corrupted input. The noising scheme includes text shuffling and spans MLM in which multiple tokens are replaced with a single masked token. The final model should be an extension of both BERT \cite{devlin2018bert} and GPT \cite{brown2020language} as it can perform discriminative and generative tasks.

\textbf{XLNet} \cite{yang2019xlnet} is a decoder model designed for generating sequences. However, due to the discriminative benefit of the encoder model used in the transformer model, the authors proposed pre-training objectives that mimic the effect of training an encoder model to enhance their decoder model.  The encoder subunit provides a better understanding of the sequence as it can process the text bi-directionally. Therefore, a modification of the MLM objective called permutation language modeling (PLM) was proposed. The key idea behind XLNet is to model the probability of a sequence of tokens by factorizing it in a way that allows all permutations to be considered. This is achieved by training the model to predict the probability of the masked token given all the tokens that appear before it in a random permutation of the input sequence.

\subsection{Molecular transformer models}

This section will outline the articles that have employed a sequence-based chemical language transformer model for molecular property prediction (Table \ref{tab:articles}). These articles will be the ones to be analyzed in the coming sections.  

\begin{table}[]
\makebox[\textwidth]{\begin{tabular}{llllll} \hline
\textbf{Architecture}            & \textbf{Base Model}                                       & \textbf{Model Name}                        & \textbf{Peer Reviewed} & \textbf{Date} & \textbf{Country} \\ \hline
\multirow{4}{*}{Encoder-Decoder} & \multirow{3}{*}{Transformer \cite{vaswani2017attention}  } & ST \cite{honda2019smiles}                  & -                 & 2019       & JPN, USA         \\
                                 &                                                            & Transformer-CNN \cite{karpov2020transformer} & \checkmark      & 2020       & DEU, CHE       \\
                                 &                                                            & X-Mol \cite{xue2022x}                      & \checkmark        & 2022       & CHN              \\ \cline{2-6}
                                 & BART \cite{lewis2020bart}                                  & ChemFormer \cite{irwin2022chemformer}      & \checkmark        & 2022       & SWE              \\ \hline
\multirow{10}{*}{Encoder}        & \multirow{5}{*}{BERT \cite{devlin2018bert}}               & SMILES-BERT \cite{wang2019smiles}          & \checkmark        & 2019       & USA                                               \\
                                 &                                                           & MAT \cite{maziarka2020molecule}            & -                 & 2020       & POL, USA         \\
                                 &                                                           & MolBERT \cite{fabian2020molecular}         & -                 & 2020       & LUX              \\
                                 &                                                           & Mol-BERT \cite{li2021mol}                  & \checkmark        & 2021       & CHN              \\
                                 &                                                           & Chen et al. \cite{chen2021extracting}     & \checkmark        & 2021       & CHN, USA         \\                                 
                                 &                                                           & K-BERT \cite{wu2022knowledge}              & \checkmark        & 2022       & CHN              \\
                                 &                                                           & FP-BERT \cite{wen2022fingerprints}         & \checkmark        & 2022       & CHN              \\ 
                                 &                                                           & MolFormer \cite{ross2022large}             & \checkmark        & 2022       & USA              \\\cline{2-6}
                                 & \multirow{3}{*}{RoBERTa \cite{liu2019roberta}}            & ChemBERTa \cite{chithrananda2020chemberta} & -                 & 2020       & USA              \\
                                 &                                                           & ChemBERTa-2 \cite{ahmad2022chemberta}      & -                 & 2022       & USA              \\
                                 &                                                            & SELFormer \cite{yuksel2023selformer}       & \checkmark        & 2023       & TUR              \\ \hline
Decoder                          & XLNet \cite{yang2019xlnet}                                 & RT \cite{born2023regression}               & \checkmark        & 2023       & CHE              \\ \hline
\end{tabular}}
\caption{A list of the reviewed articles that used a sequence-based transformer model for molecular property prediction. An Encoder-Decoder model is suitable for both discriminative and generative tasks, an encoder-only model is more suitable for discriminative tasks, and a decoder-only model is more suitable for generative tasks. The model name is adopted as stated in the corresponding article.}
\label{tab:articles}
\end{table}

\textbf{SMILES-Transformers (ST)} \cite{honda2019smiles}, \textbf{ChemFormer} \cite{irwin2022chemformer}, \textbf{Transformer-CNN} \cite{karpov2020transformer}, and \textbf{X-Mol} \cite{xue2022x} are encoder-decoder models. This entails that these models are suitable for both discriminative and generative tasks. \textbf{ST} \cite{honda2019smiles} employed the original transformer \cite{vaswani2017attention} model as an autoencoder and utilized the output of the encoder model as an embedding for the molecule to use in MPP. \textbf{ChemFormer} \cite{irwin2022chemformer} adopted the BART \cite{lewis2020bart} model, which employs the transformer model as a denoising autoencoder. The authors utilized the non-uniqueness feature of the SMILES language (i.e., one molecule can be represented by different SMILES) as a noising scheme called augmentation. This objective inputs a non-canonical sequence to the model, and the corresponding canonical sequence is constructed. The authors also experimented with using the span MLM objective proposed in BART, as well as a combination of both augmentation and span MLM. \textbf{X-Mol} \cite{xue2022x} and \textbf{Transformer-CNN} \cite{karpov2020transformer} also utilized the non-uniqueness feature of the SMILES to train a transformer model to decode an equivalent SMILES of a given input SMILES. However, instead of independent encoder-decoder subunits as presented in the transformer \cite{vaswani2017attention} model and used by the Transformer-CNN \cite{karpov2020transformer}, the X-Mol \cite{xue2022x} concatenated both the input SMILES and the desired output SMILES in the same training space. They fully masked the output SMILES and allowed the model to encode the input SMILES bi-directionally while decoding the masked output SMILES uni-directionally.

Instead of using the encoder-decoder architecture, most papers utilized only the encoder model, as it can perform discriminative tasks independently of the decoder. \textbf{SMILES-BERT} \cite{wang2019smiles} is one of the first papers to employ the encoder model for MPP by using BERT \cite{devlin2018bert}. As mentioned above, the BERT model employs two pre-training objectives, MLM and NSP. However, the authors utilized the MLM objective only, arguing that the NSP objective is unsuitable for molecular data. They argued that if molecules were equated to sentences, different molecules would not follow a consecutive order to each other, hence, no need for the NSP objective. 

\textbf{MolBERT} \cite{fabian2020molecular} also utilized the MLM objective only. However, the authors proposed two additional objectives specific to the molecules and the chemical language, introducing inductive bias to the model. These two objectives are PhysChemPred and SMILES-EQ. The former objective incorporated the prediction of $\sim 200$ physico-chemical properties of a molecule, while SMILES-EQ attempted to enhance the model's comprehension of the non-uniqueness of the SMILES language by asking it to identify the different but equivalent SMILES. \textbf{K-BERT}\cite{wu2022knowledge} attempted a similar inductive bias approach as they trained their model using three molecule-specific objectives, namely, atom feature prediction, molecule feature prediction, and contrastive learning (CL). For atom feature prediction, features like degree, hybridization, aromaticity, etc. were predicted for each heavy atom. The molecule-level feature prediction objective was performed by predicting the corresponding molecular access system (MACCS) \cite{durant2002reoptimization} structural key. Finally, for the non-uniqueness of the SMILES language, CL, a machine-learning technique, was used to enhance the representation space as it aims to maximize the similarity between positive pairs (i.e., different SMILES of the same molecule) and minimize the similarity between negative pairs (i.e., SMILES of different molecules). \textbf{Molecule Attention Transformer (MAT)} is another model to use the inductive bias, however, the authors decided to do so by augmenting the self-attention approach proposed in the transformer model with structural information about each atom such as inter-atomic distances and adjacency in the molecular graph. The output of the now-called molecule self-attention layer is a weighted sum of the self-attention, distance information, and adjacency matrix for each atom. Furthermore, the authors replaced the text-based representation with a simple list of atoms composing a molecule. To this end, each atom in the list is embedded as a 26 vector that indicates the type of the atom and some structural information like the number of neighboring atoms, being a part of a ring, etc. 

Moving away from inductive bias, \textbf{Chen et al.} \cite{chen2021extracting} investigated the relationship between the pre-training and downstream data sets (i), \textbf{Mol-BERT} \cite{li2021mol} and \textbf{FP-BERT} \cite{wen2022fingerprints} focused on improving the tokenization method to yield more informative tokens (ii). In contrast, \textbf{MolFormer} \cite{ross2022large} modified the implementation of the model to train more efficiently and scalably (iii). Starting with the data sets (i), \textbf{Chen et al.} \cite{chen2021extracting} trained three models with different pre-training data set combinations. They then implemented a decision module to determine the most suitable model for a downstream data set based on the similarity in distributions between the pre-training and downstream data sets. Moving to explore the input of the model (ii), instead of using the text-based SMILES representation as is, \textbf{Mol-BERT} \cite{li2021mol} and \textbf{FP-BERT} \cite{wen2022fingerprints} followed a previously proposed approach of treating the molecule as a sentence in an analogy to natural languages \cite{jaeger2018mol2vec}. In this setup, the composing units of the molecule would be analogous to the words in a sentence. These units were computed with Morgan fingerprints (or ECFP) by hashing the substructures of radius 1 for each atom beside the atom itself. Therefore, the molecules are represented by concatenating the hashed values of their atoms and atom environments. Ending with optimizing the implementation (iii), \textbf{MolFormer} \cite{ross2022large} proposed to implement linear attention and rotary positional embedding. On one hand, the linear attention approach proposed by Katharopoulos et al. \cite{katharopoulos2020transformers} was used to counter the original quadratic scaling of the attention mechanism. On the other hand, the rotary encoding proposed in RoFormer \cite{su2021roformer} was meant to provide flexibility for the input sequence length. 

Besides BERT \cite{devlin2018bert}, RoBERTa \cite{liu2019roberta} was also used as a base model for \textbf{ChemBERTa} \cite{chithrananda2020chemberta}, \textbf{ChemBERTa -2} \cite{ahmad2022chemberta}, and \textbf{SELFormer} \cite{yuksel2023selformer}. ChemBERTa-2 is an extension of ChemBERTa, and both aim to explore different options regarding training decisions. \textbf{ChemBERTa} \cite{chithrananda2020chemberta} explored decisions like the effect of pre-training data set size, the effect of the input molecular language (i.e., SMILES vs SELFIES), as well as multiple hyperparameters tuning for the model's architecture. \textbf{ChemBERTa-2} \cite{ahmad2022chemberta} additionally experimented with the choice of the pre-training objectives by comparing the traditional MLM objective to Multi-task regression (MTR). The latter objective is the same as the PhysChemPred objective used in MolBERT \cite{fabian2020molecular}. \textbf{SELFormer} \cite{yuksel2023selformer}, on the other hand, is aimed at showcasing the performance of the transformer models using the SELFIES language.

The decoder model is mainly used for generative tasks. However, the authors of \textbf{Regression Transformers (RT)} \cite{born2023regression} managed to build a generative model that is capable of performing MPP. The \textbf{RT} \cite{born2023regression} model is based on XLNet \cite{yang2019xlnet} and is aimed to perform molecular optimization by generating molecules from some given molecular properties. The authors proposed to do so by prepending the numerical values of the properties as tokens to the input sequence and using the cross-entropy loss to predict the most probable next tokens to generate the desired molecule. To perform MPP, the authors quantified some regression data sets by masking the prepended numerical tokens of a molecule and allowing the model to decode these values using only the cross-entropy loss.

To this end, we have introduced the models that use sequence-based chemical languages to train a transformer model or one of its variants to predict molecular properties. We have provided a comprehensive list to the best of our knowledge, but it might not be a complete one. 

\section{Data sets used for training molecular transformers} 

The self-supervised learning scheme with transformers typically requires two types of data sets, which are introduced in the following subsections. Pre-training data sets are usually large unlabeled data collections used for extracting general features of molecules. Downstream data sets are smaller labeled data sets used for fine-tuning a transformer model for a specific domain-relevant task, like predicting a molecule's solubility.    

\subsection{Pre-training data sets} 
\label{sec:pre-training-data}

Data sets for pre-training transformers are mostly large chemical databases maintained and curated by institutions and organizations.  
We will first introduce the main large-scale data sets ZINC, ChEMBL, and PubChem, followed by benchmark subsets drawn from those.

\paragraph{ZINC} is an openly accessible database that gathers commercially available compounds with a focus on "drug-like" molecules \cite{irwin2005zinc, sterling2015zinc, irwin2020zinc20, tingle2023zinc}. Many molecules are annotated with physico-chemical properties, 3D conformers, and pharmacological and biological information\cite{sterling2015zinc, irwin2020zinc20}. From the start, ZINC focused on offering purchasable molecules for virtual screening and lead discovery \cite{irwin2005zinc, irwin2020zinc20}, leading to the prioritization of molecules with molecular weight $\leq{500}$ Dalton and calculated logP $\le{5}$ \cite{irwin2020zinc20}. The current ZINC22 database contains more than \textbf{$37$ billion molecules} coming from both in-stock and make-on-demand collections \cite{tingle2023zinc}. The ZINC database is growing rapidly by including enumerated molecule collections, such as Enamine REAL \cite{enamineREAL}. For comparison, the ZINC20 contained only around $2$ billion molecules \cite{irwin2020zinc20}.

\paragraph{ChEMBL} is a curated and openly available database providing bioactivity data focusing on molecules with drug-like properties \cite{mendez2019chembl, zdrazil2023chembl}. The current ChEMBL release v.33 contains \textbf{$2.4$ million unique molecules} and more than $20$ million bioactivity measurements \cite{zdrazil2023chembl}. Approximately $9$ million of the bioactivities were extracted manually by curators from the medicinal chemistry and pharmaceutical literature, while the other $11$ million came from deposited data sets \cite{zdrazil2023chembl}. The data widely covers the drug discovery process and comes from $1.6$ million assays measured on more than $17$ thousand targets  \cite{zdrazil2023chembl}. In addition, multiple categorizations like assay types, target types or action types are provided \cite{zdrazil2023chembl}. 

\paragraph{PubChem} is a public chemical database aggregating various information on mostly small molecules from multiple sources \cite{kim2023pubchem, kim2016pubchem}. The database is maintained by the National Center for Biotechnology Information (NCBI) and started with the aim to provide biological high-throughput test results of small molecules \cite{wang2009pubchem, kim2023pubchem}. Today, PubChem provides broad annotations from more than $870$ sources grouped into data collections containing information about substances, bioassays, protein targets, genes, pathways, cell lines, taxonomy, and patents \cite{kim2023pubchem}.
Currently, more than \textbf{$111$ million unique molecules} are deposited in PubChem \cite{kim2023pubchem}.
%PubChem's data comes from various organizations and individual contributors \cite{kim2016pubchem}. 
Note that, for example, the bioactivity data of ChEMBL is incorporated into PubChem, and parts of PubChem's bioactivity data are also in ChEMBL \cite{kim2016pubchem}.   

\paragraph{Subsets of these major data sources:}
Only some approaches report using complete databases for pre-training. For example, MolFormer \cite{ross2022large} used the complete PubChem, which they combined with $1$ billion molecules from ZINC. In addition, K-BERT \cite{wu2022knowledge} and SELFormer \cite{yuksel2023selformer} seem to have been pre-trained with the full ChEMBL.
However, most of the transformer approaches were pre-trained on subsets of those data sets created by random sampling, filtering, or extraction processes without further specification.
MolBERT \cite{fabian2020molecular} used, for example, the GuacaMol data set \cite{brown2019guacamol}, a subset of ChEMBL containing $1.6$ million molecules filtered and standardized for evaluating \textit{de novo} molecular design methods. Li et al. used in Mol-BERT \cite{li2021mol} a combination of ZINC and ChEMBL filtered based on physico-chemical properties and chemical elements yielding $4$ million molecules. FP-BERT \cite{wen2022fingerprints} used randomly sampled compounds from the E15 data set, some not further specified subset \cite{babuji2020targeting} of the Enamine REAL data set \cite{enamineREAL} containing about $15.5$ million molecules. The other data sets either used random sampling or did not mention their subset extraction process.

\subsection{Downstream data sets} \label{downstream data}

Data sets used for downstream applications are usually labeled and significantly smaller than the pre-training data sets. They often come from individual publications and were assembled in benchmark suites, such as MoleculeNet \cite{wu2018moleculenet} or Therapeutics Data Commons (TDC)  \cite{huang2021therapeutics}. %collect these data sets and make them accessible as data set collections. 
Supplementary Tables \ref{tab:data_calssification} and \ref{tab:data_regression} show which downstream data sets have been evaluated by which model. Table \ref{tab:downstream_datasets} lists the most frequently used data sets for fine-tuning and evaluating transformers in the literature. The data sets listed are the versions from MoleculeNet, which are used exclusively in the reviewed papers.

\begin{table}
\centering
\caption{Overview of data sets used for evaluating transformers for property prediction. Footnotes: $^1$ data set is included in the MoleculeNet collection. $^2$ data set is included in the TDC collection. $^3$ These percentages correspond to the minimum and maximum of all tasks. Detailed percentages for each task can be found in Supplementary tables \ref{tab:clintox_pos_pct}, \ref{tab:sider_pos_pct}, and \ref{tab:tox21_pos_pct} for ClinTox, SIDER, and Tox21, respectively.}
\label{tab:downstream_datasets}

\begin{tabular}{cccccc}
\hline
Data set & Number of & \multicolumn{2}{c}{Tasks} & \% Positive & Application \\
 & Molecules & Classif. & Regr. & Class & Area\\
\hline
\multicolumn{1}{c}{\textbf{physical chemistry}}& & \\
ESOL$^1$                              & 1128                   & 0                       & 1                   &                   -   & e.g., drug discovery, \\
FreeSolv$^{1,2}$                      & 642                    & 0                       & 1                   &               -       &         formulation,   \\
Lipophilicity$^{1,2}$                 & 4200                   & 0                       & 1                   &                  -    &         process design  \\
\\
\multicolumn{1}{c}{\textbf{physiology}} & &\\
BBBP$^{1,2}$                          & 2050                   & 1                       & 0                   & 76\%                 & ADME (distribution)                                                  \\
ClinTox$^{1,2}$                       & 1484                   & 2                       & 0                   & 8 - 94\%($^3$)                 & toxicity                                                             \\
SIDER$^1$                             & 1427                   & 27                      & 0                   & 2 - 92\%($^3$)                & drug side effects                                                    \\
Tox21$^{1,2}$                         & 7831                   & 12                      & 0                   & 3 - 16\%($^3$)                  & toxicity                                                             \\
 %                                       &                        &                         &               %     &                      &                                                                      \\
 \\
\multicolumn{1}{c}{\textbf{biophysics}}& & \\
BACE$^1$                              & 1513                   & 1                       & 1                   & 46\%                 & e.g. target interaction                                              \\
HIV$^{1,2}$                           & 41127                  & 1                       & 0                   & 4\%                  &  \\ \hline                                                                   
\end{tabular}

\end{table}

% physical chemistry data sets
The three downstream data sets in the physical chemistry category listed in Table \ref{tab:downstream_datasets} cover a compound's solvation and solubility-related physico-chemical properties. These are key physical properties for applications in various fields like drug discovery and development, formulation, chemical process design, etc. \cite{nicholls2008predicting, boobier2020machine}. These physico-chemical properties influence multiple other molecular properties. For example, a molecule's solubility is a driving factor for its ADME and toxicity profile \cite{bergstrom2018computational}. 
The ESOL data set, from the same-named method Estimated SOLubility (ESOL), comprises $1128$ molecules with experimentally determined aqueous solubility (log solubility mol/L) \cite{delaney2004esol}, the FreeSolv data contains hydration free energies of $642$ small molecules \cite{mobley2014freesolv} and the Lipophilicity data set provides $4200$ experimental measures of the octanol/water distribution coefficient (LogD at pH 7.4) provided by AstraZeneca \cite{hersey2015chembl}.  

% physiology data sets
The labels of the four data sets in the physiology category describe endpoints related to toxicity, side effects and distribution in organisms. The blood-brain barrier penetration (BBBP) data set \cite{martins2012bayesian} provides $2050$ labeled molecules indicating whether the molecule can pass the blood-brain barrier. The development of drugs targeting the central nervous system can be challenging since they must be able to penetrate this barrier \cite{martins2012bayesian}. Vice versa, unintended permeability might induce side effects and toxicity \cite{martins2012bayesian}.
The ClinTox data set contains $1484$ molecules with labels indicating whether the molecule failed clinical trials due to toxicity and whether it was approved by the FDA \cite{gayvert2016data, wu2018moleculenet}. 
The SIDER data set provides multiple labels for adverse reactions of $1427$ drugs\cite{kuhn2016sider}. Its $27$ endpoints represent system organ classes according to the Medical Dictionary for Regulatory Activities \cite{wu2018moleculenet}.
Lastly, the Tox21 data set contains $7831$ molecules with high-throughput screening results for a total of $12$ nuclear receptor binding and interactions with stress response pathways. The disruption of these pathways can result in various diseases such as cancer, diabetes, or Parkinson's \cite{zhao2019nuclear, fulda2010cellular}.

% biophysic data sets
The two biophysics data sets describe endpoints of small molecule target interaction. The BACE data set contains IC50 binding affinity measures extracted from literature and derived binary labels for the binding of $1513$ small molecules to the human $\beta$-secretase 1, a key therapeutic target in Alzheimer's \cite{subramanian2016computational}. In contrast, the HIV data set contains $41127$ molecules categorized by their anti-HIV activity. Experts categorized molecules based on experimentally determined EC50 and IC50 measurements from cell-based screening, indicating a protecting effect \cite{nci_wiki_hiv}. The MoleculeNet collection \cite{wu2018moleculenet} provides a binarized version of the HIV data set's categories.    

\subsection{The current SOTA performance for some downstream data sets}
\label{sota}

In this section, we present an overview of the performance of the different transformer models (Table \ref{tab:articles}) on the downstream data sets (Table \ref{tab:downstream_datasets}).
However, through our literature study, we encountered several obstacles that rendered a detailed and fair comparison infeasible. Tables \ref{tab:splits} and \ref{tab:evaluation} collect information about the different data splits and performance evaluations, respectively. 

Most articles used scaffold splitting for classification and random splitting for regression tasks (Table \ref{tab:splits}). However, several articles evaluated classification tasks with random splitting instead. Random splitting is generally a less preferred method as it risks data leakage to the test set and, therefore, may overestimate a model's performance \cite{yang2019analyzing, deng2023systematic}. 
Importantly, the test set is not guaranteed to be unified across the models using the same split method (e.g. scaffold splitting), as the data sets were split independently in each publication. Moreover, while a train/valid/test split of 80/10/10 was most predominately, some papers used different splitting ratios. 
Even if the same splitting ratio is used, if the molecules' identifiers in the individual subsets per run are not given, reported numbers can not be directly compared since one can not guarantee that the same underlying test sets were evaluated. 
Furthermore, the different articles performed different numbers of cross-validation repetitions. Therefore, some analyses are more robust than others.
Considering the performance evaluation metrics, mostly ROC-AUC and RMSE were reported for classification and regression, respectively; except for some models (see Supplementary Table \ref{tab:evaluation}). The studies also differ in the evaluation detail; while some authors report statistical analysis, others only provide averaged values. Thus, the comparison is further hindered since a single averaged number can not fairly be compared to, e.g., values with reported standard deviations. \cite{weissgerber2017data, midway2020principles}. 

\begin{table}[]
\makebox[\textwidth]{\begin{tabular}{llllll}
\hline
\textbf{Model Name}                        & \multicolumn{3}{c}{\textbf{Split Method}}                   & \textbf{Split Ratio} & \textbf{\# Rep.} \\ \hline
\textbf{}                                  & \textbf{Classification} & \textbf{Regression} & \textbf{By} & \textbf{}            & \textbf{}           \\ \hline
MolBERT \cite{fabian2020molecular}         & Scaffold                & Random              & ChemBench   & 80/5/15              & NA                  \\
SELFormer \cite{yuksel2023selformer}       & Scaffold, Random        & Random              & ChemProb    & 80/10/10             & NA                  \\ 
ST \cite{honda2019smiles}                  & Scaffold                & Random              & Self ($^1$) & 80/10/10             & 20                  \\
MAT \cite{maziarka2020molecule}            & Scaffold                & Random              & Self ($^1$) & 80/10/10             & 6                   \\
MolFormer \cite{ross2022large}             & Scaffold                & Random              & Self($^1$)  & NA                   & NA                  \\
FP-BERT \cite{wen2022fingerprints}         & Scaffold                & Random              & DeepChem    & 80/10/10             & 5                   \\
ChemBERTa-2 \cite{ahmad2022chemberta}      & Scaffold                & Random              & DeepChem    & 80/10/10             & NA                  \\
ChemBERTa \cite{chithrananda2020chemberta} & Scaffold                & -                   & DeepChem    & 80/10/10             & NA                  \\
Mol-BERT \cite{li2021mol}                  & Scaffold                & -                   & Self        & 80/10/10             & 5                   \\
SMILES-BERT \cite{wang2019smiles}          & Random                  & -                   & Self($^1$)  & 80/10/10             & NA                  \\
K-BERT \cite{wu2022knowledge}              & Random                  & -                   & Self($^1$)  & 80/10/10             & NA                  \\
Chen et al. \cite{chen2021extracting}     & Random                  & Random              & Self($^1$)  & 80/10/10             & 50                    \\
Transformer-CNN \cite{karpov2020transformer}     & Random ($^1$)     & Random ($^1$)       & Self($^1$) &  NA                   & 5                    \\
X-Mol \cite{xue2022x}                      & Random                  & Random              & Self($^1$)  & NA                   & 20                  \\
ChemFormer \cite{irwin2022chemformer}      & -                       & Random              & Self($^1$)  & 75/10/15             & 20                  \\
RT \cite{born2023regression}               & -                       & Random              & NA          & NA/NA/15             & 3                   \\
\hline

\end{tabular}}
\caption{Evaluation data set pre-processing made by the reviewed articles. The "split method" specifies the method used to split the data and by whom was this split performed. The "split ratio" column specifies the train/valid/test ratios. The "\# Rep." specifies the number of repetitions of the evaluation process. NA corresponds to unreported information by the authors. Footnotes: $^1$ This information was inferred for the lack of contradicting information. }
\label{tab:splits}
\end{table}

\begin{figure}
  \includegraphics[width=1.0\linewidth]{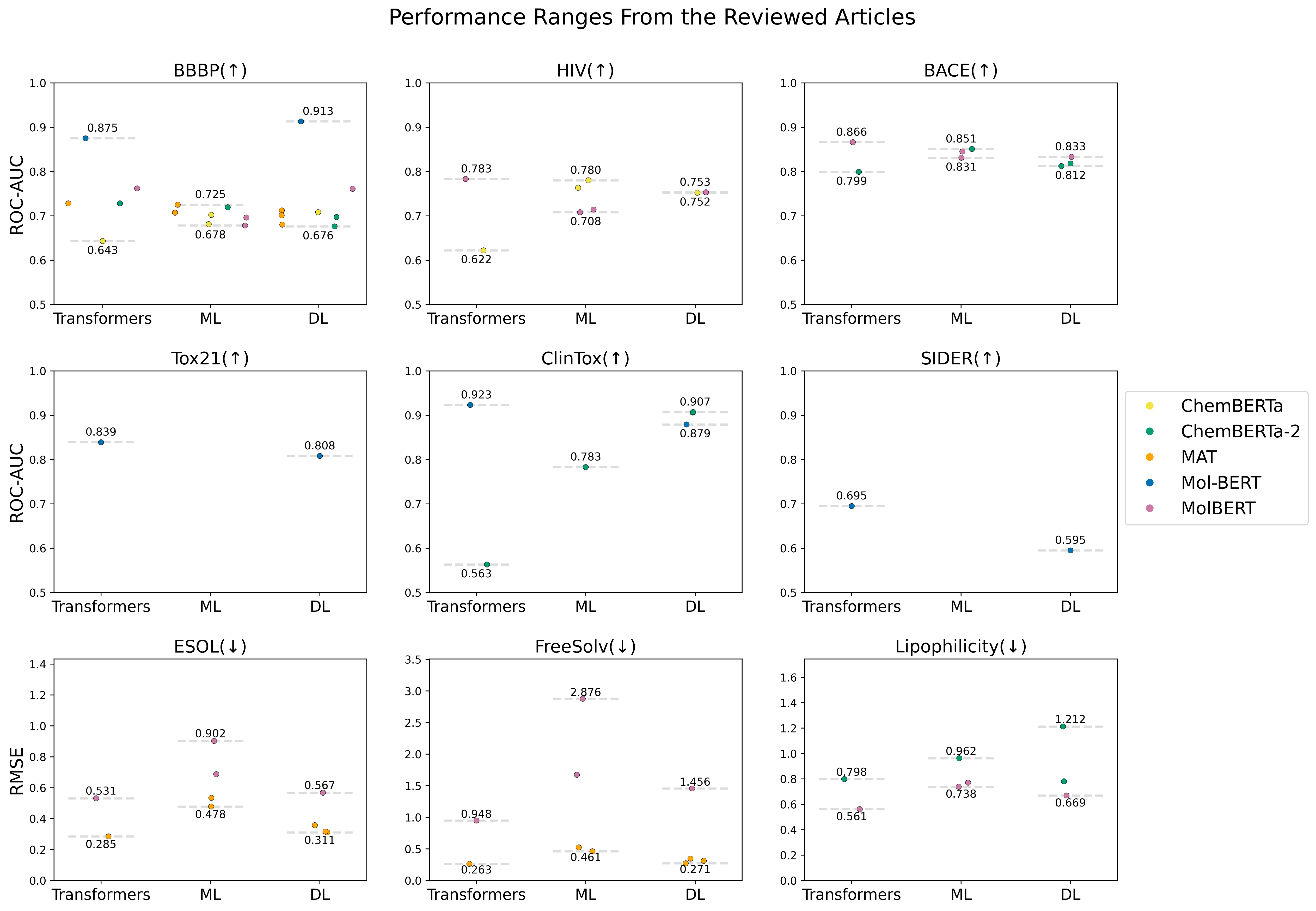}
  \caption{A comparison between the ROC-AUC and RMSE ranges for the reviewed articles, some classical machine learning (ML) algorithms, and some deep learning (DL) models. Scaffold splitting was used for the classification data sets and random splitting was used for the regression data sets by all the models. Only models with the same color were tested on the same test sets. 
  The values for the transformers models span some models in Table \ref{tab:articles}, which are shown on the side legend. The reported classical ML models are RF and SVM. The DL models span graph-based and different DNN models like D-MPNN, Weave, etc. The values for classical ML and DL categories are obtained from the comparisons done by the transformer models shown on the side legend. Supplementary Table \ref{tab:models_comparable_comparisons} shows which classical ML and DL models were used for comparison by each transformer model. Data and code used to generate this figure can be found in \href{https://github.com/volkamerlab/Transformers4MPP_review/tree/main}{our GitHub repo}.}
  \label{fig:performance_honest}
\end{figure}

To show the potential of molecular transformers, the different published models compared their performance against other classical machine learning (ML) and deep learning (DL) models. We collected the values from the respective publications and plotted the current state-of-the-art (SOTA) for each mentioned data set (Supplementary Figure \ref{fig:performance}). However, several works used the reported performance values rather than re-running the compared models on their specified test set, which hinders fair comparison, as stated earlier. Therefore, for the least-biased comparison, Figure \ref{fig:performance_honest} shows performance values across models guaranteed to be calculated on the same test sets.
The figure shows promising but varying trends for the different transformer models. For example, the Mol-BERT \cite{li2021mol} model from Li et al. reports a lower ROC-AUC compared to its corresponding DL model (MPNN \cite{gilmer2017neural}) for the BBBP data set ($\sim$ 0.88 and 0.91, respectively). However, the MolBERT \cite{fabian2020molecular} model from Fabian et al. performed better than its corresponding DL model (CDDD \cite{winter2019learning}) with an ROC-AUC of $\sim$ 0.86 and 0.76, respectively, as well as its corresponding SVM model with RDKit descriptors (ROC-AUC of $\sim$ 0.86 and 0.70, respectively) for the BBBP data set. The same trend was observed for the HIV data set, as ChemBERTa \cite{chithrananda2020chemberta} performed worse than its corresponding ML models, while MolBERT \cite{fabian2020molecular} performed better than its corresponding ML ($\sim 0.08$ $\triangle ROC-AUC$) and DL ($\sim 0.03$ $\triangle ROC-AUC$) models. 
For the toxicity datasets ( \ref{fig:performance_honest} middle row), fewer comparable models were available. However, the only observation for the SIDER dataset, for example, shows a prominent difference as Mol-BERT performed $\sim 0.1$ better ROC-AUC than its corresponding MPNN \cite{gilmer2017neural} model. 
For regression, MAT \cite{maziarka2020molecule} and MolBERT \cite{fabian2020molecular} showed improved performance compared to their corresponding ML and DL models for the three data sets (ESOL, FreeSolv, and LipoPhilicity). For example, MAT \cite{maziarka2020molecule} performed $\sim 0.2$ $\triangle RMSE$ better than an SVM model and $\sim 0.03$ $\triangle RMSE$ better than the Weave \cite{kearnes2016molecular} model for the ESOL dataset.      

% different models have very different decisions.
Overall, the transformer model shows promising results in MPP for many tasks. 
However, this improvement is not consistent across the different data sets or models. The varying performance across models can be explored in terms of the choices made during the building, pre-training, and fine-tuning of the transformer model. For example, the ST\cite{honda2019smiles} model is an encoder-decoder that was trained on $\sim 900K$ molecules, while the MolFormer \cite{ross2022large} model is an encoder-only that is trained on 1.1B molecules. Most of the models used SMILES as a representation language. But some models were trained using SELFIES \cite{yuksel2023selformer, chithrananda2020chemberta, born2023regression} or circular fingerprints \cite{li2021mol, wen2022fingerprints}. The Mol-BERT \cite{li2021mol} model consisted of $\sim 11M$ parameters, while the MAT \cite{maziarka2020molecule} model used $> 100M$ parameters. Most of the models were trained with domain-agnostic objectives like MLM. However, models like MolBERT \cite{fabian2020molecular}, K-BERT \cite{wu2022knowledge}, ChemBERTa-2 \cite{ahmad2022chemberta}, and MAT \cite{maziarka2020molecule} incorporated domain-relevant objectives. Consequently, questions arise, such as which choices have been explored so far, which approaches have been more suitable for molecular language models, and which still need more analysis. Trying to answer these questions will be the focus of the remainder of this review. 

\section{The decisions to consider when implementing a transformer model for MPP} 
\label{sec-decisions}

\begin{figure}
  \includegraphics[width=1.0\linewidth]{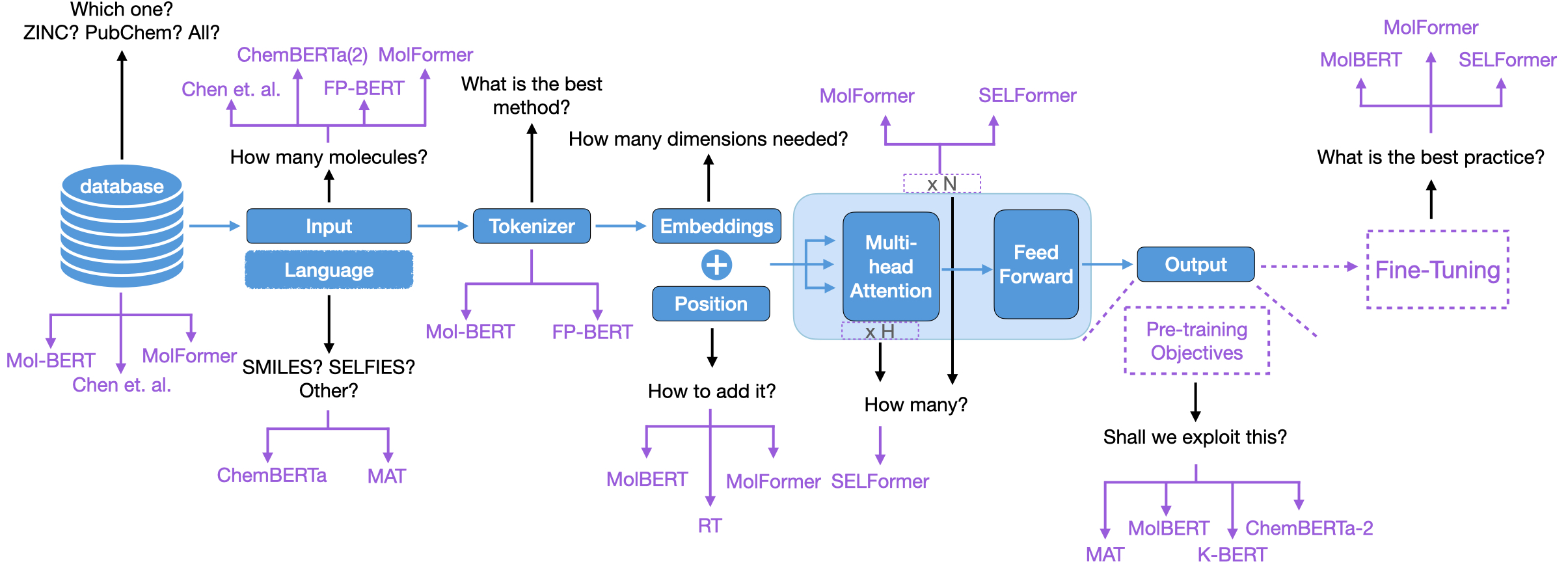}
  \caption{An overview of the individual components of the transformer model and which decisions were explored by which article. Black text represents the set of questions that were asked in the following subsections.}
  \label{fig:decisions}
\end{figure}

In this section, we will further explore the individual decisions that must be made during training and fine-tuning a transformer model for MPP. 
Figure \ref{fig:decisions} summarizes the individual components of the transformer model that the reviewed models investigated. We raise a question for each component, and the subsections below work along with these questions.

\subsection{Which database to use for pre-training, and how many molecules should it contain?}

The goal of the pre-training data set is to provide a generic representation of the molecules without the need for labeled information. On the one hand, it has been shown for NLP that selecting the most appropriate data set in terms of diversity or relatedness to the downstream tasks can help the model with faster and better learning and less carbon footprint \cite{albalak2024survey}. On the other hand, the current trend in NLP, for models like GPT, is moving toward increasing the data set size as much as possible \cite{radford2019language}. In addition, several studies suggest that the needed data set size depends on the number of parameters used \cite{kaplan2020scaling, hoffmann2022training}. In this section, we examine how the current chemical language models approached both the data set selection and size for their models.

Molecular databases (see Section \ref{sec:pre-training-data}) are assembled and curated for different purposes. The contained molecules can differ and represent different chemical spaces.
Table \ref{tab:pre-training_datasets} lists transformers from the literature with the databases they were pre-trained on. Some approaches combined molecules from different databases and actively compared the influence of different data sources on prediction performance. 

\begin{table}[]
\makebox[\textwidth]{\begin{tabular}{llll} 
\hline
\textbf{Model Name}     & \multicolumn{2}{c}{\textbf{Pre-training data set}} & \textbf{Language} \\ \hline
                                                  & \textbf{Database} & \textbf{Size ($\sim$)}            \\ \hline
ST \cite{honda2019smiles}                         & ChEMBL            & 900K                 & SMILES           \\ 
RT \cite{born2023regression}                      & ChEMBL            & 1.4M                 & SELFIES          \\
MolBERT \cite{fabian2020molecular}                & ChEMBL (GuacaMol) & 1.6M                 & SMILES           \\
Transformer-CNN \cite{karpov2020transformer}      & ChEMBL            & 1.8M                 & SMILES           \\
K-BERT \cite{wu2022knowledge}                     & ChEMBL            & 1.8M                 & SMILES           \\
SELFormer \cite{yuksel2023selformer}              & ChEMBL            & 2M                   & SELFIES          \\
ChemBERTa \cite{chithrananda2020chemberta}        & PubChem           & 100K, 250K, 1M, 10M  & SMILES, SELFIES  \\
ChemBERTa-2 \cite{ahmad2022chemberta}             & PubChem           & 5M, 10M, 77M         & SMILES           \\
MAT \cite{maziarka2020molecule}                   & ZINC              & 2M                   & List of atoms    \\ 
SMILES-BERT \cite{wang2019smiles}                 & ZINC              & 19M                  & SMILES           \\
ChemFormer \cite{irwin2022chemformer}             & ZINC              & 100M                 & SMILES           \\
X-Mol \cite{xue2022x}                             & ZINC              & 1.1B                 & SMILES           \\
Mol-BERT \cite{li2021mol}                         & ZINC + ChEMBL     & 4M                   & Circular FPs     \\
MolFormer \cite{ross2022large}                    & ZINC + PubChem    & 1.1B                 & SMILES           \\
Chen et al. \cite{chen2021extracting}             & C, CP, and CPZ    & 2M, 103M, 775M       & SMILES           \\
FP-BERT \cite{wen2022fingerprints}                & E15               & 2M, 10M              & Circular FPs              \\ 
\hline
\end{tabular}}
\caption{The database and the number of molecules used for pre-training each transformer model. The "Database" column shows the source of the pre-training data set. The "Size" column shows the approximate number of molecules used for training each model. The "Language" column shows the molecular representation language used by the corresponding model. FP = Fingerprints, it's Morgan fingerprints for Mol-BERT \cite{li2021mol} and ECFP4 for FP-BERT \cite{wen2022fingerprints}. C = ChEMBL, CP = ChEMBL + PubChem, and CPZ = ChEMBL + PubChem + ZINC.}
\label{tab:pre-training_datasets}
\end{table}

\begin{figure}
  \includegraphics[width=1.0\linewidth]{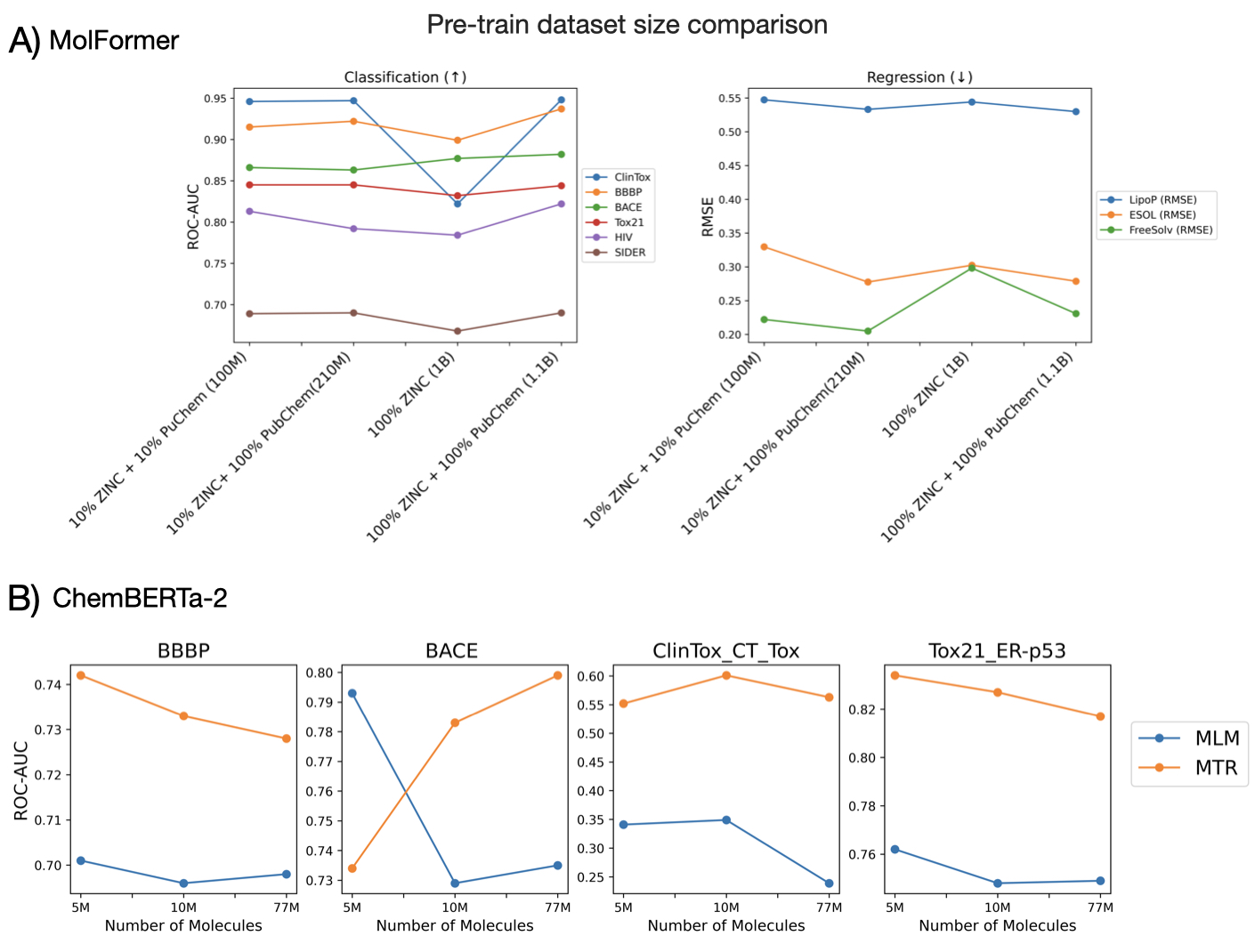}
  \caption{Comparison between the performance and pre-training data set size by \textbf{A)} MolFormer \cite{ross2022large} and \textbf{B)} ChemBERTa-2 \cite{ahmad2022chemberta}. The models are sorted in ascending order based on the size of the pre-training data set. $^*$ MLM = Masked language modeling. MTR = Multi-task regression. Data and code used to generate this figure can be found in \href{https://github.com/volkamerlab/Transformers4MPP_review/tree/main}{our GitHub repo}.}
  \label{fig:size_performance}
\end{figure}

Chen et al. \cite{chen2021extracting} trained three models with different pre-trained data set combinations, namely, ChEMBL (C), ChEMBL + PubChem (CP), and ChEMBL + PubChem + ZINC (CPZ), containing roughly 2, 103, and 775 million compounds, respectively. The reported performance on five regression data sets for the three models is comparable, i.e., average performance values with overlapping standard errors. For example, for ESOL, an $R^2$ of $0.925 \pm 0.01$, $0.917 \pm 0.012$, and $ 0.915 \pm 0.01$ were reported for C, CP, and CPZ models, respectively. A slightly higher performance trend can be observed for the model pre-trained with only the \textit{smaller} ChEMBL data set.  
MolFormer \cite{ross2022large} was also trained on a combination of two databases, PubChem and ZINC, containing a total of 1.1 billion molecules. 
The authors reported performance on four additional models trained with weighted combinations of both databases, namely, $10\%$ ZINC + $10\%$ PubChem ($\sim 100$ M molecules), $10\%$ ZINC + $100\%$ PubChem ($\sim 210$ M molecules), and $100\%$ ZINC ($\sim 1$ B molecules). Figure \ref{fig:size_performance}(A) shows that the model trained with the \textit{smaller} data set ($10\%$ ZINC + $10\%$ PubChem) is performing very close to the model trained with ten times the pre-training data set size on the different downstream data sets. For both models, average ROC-AUC values of $\sim 0.85$ were reported for the six classification tasks. In comparison, for three regression tasks, the models trained on the larger data set slightly outperformed those with a smaller data set, with RMSE values of $\sim 0.35$ and $0.37$, respectively. Moreover, the model trained with the second largest data set (100\% ZINC) performed worst for the examined downstream data sets (the average difference in ROC-AUC and RMSE is $\sim 0.04$ and $\sim 0.03$, respectively). Figure \ref{fig:size_performance} also shows that the lower performance by training solely on ZINC was not observed when combining only $10\%$ of ZINC with $10\%$ of PubChem. The authors explained the difference in performance by the low vocabulary size and short sequence length of ZINC compared to the other data set combinations. 

Other models like ChemBERTa \cite{chithrananda2020chemberta}, ChemBERTa-2 \cite{ahmad2022chemberta}, and FP-BERT \cite{wen2022fingerprints} conducted dedicated experiments to analyze the pre-training data set size. ChemBERTa \cite{chithrananda2020chemberta} trains four models with 100K, 250K, 1M, and 10M molecules from PubChem. They observe that the increase in performance from training on 100K molecules to 10M molecules for three classification tasks is $\sim 0.11$ ROC-AUC and $\sim 0.06$ PRC-AUC on average. ChemBERTa-2 \cite{ahmad2022chemberta}, however, does not come to the same conclusion as shown in figure \ref{fig:size_performance}(B). The authors also trained three models with 5M, 10M, and 7M molecules from PubChem. Additionally, two models were trained using different pre-training objectives, MLM and MTR. Figure \ref{fig:size_performance}(B) shows that increasing the data set size does not always correspond to increased performance for the tested classification tasks. For example, the model trained on the smallest data set using the MLM objective performs almost the same as the model trained on the largest data set for BBBP ($\sim 0.70$ ROC-AUC). In contrast, the same model trained on 5M molecules performs $\sim 0.01$ ROC-AUC better than the model trained on 77M molecules when using the MTR objective. Finally, the FP-BERT \cite{wen2022fingerprints} model trained on 10M molecules showed an increase in ROC-AUC of $\sim 0.01$ and $\sim 0.02$ for BBBP and HIV data sets, respectively, compared to the model trained on 2M molecules from the E15 data set. However, the same trend did not hold for regression as the model trained on 10M molecules from the same data set performed almost the same as the model trained on 2M molecules for ESOL and FreeSolv ($\sim 0.67$ and $\sim 1.1$ RMSE, respectively, for both models), while the model trained on the smaller data set performed $\sim 0.01$ RMSE better for Lipophilicity.

Drawing a clear conclusion from the provided data is difficult. The data set selection and size need further systematic analysis to assess the benefit of scaling. Such systematic analysis has been performed in NLP in publications like Kaplan et al. \cite{kaplan2020scaling} and Hoffmann et al. \cite{hoffmann2022training}. It would be beneficial to perform a similar analysis for molecular language models. Additionally, the selection of the data set can also affect the performance when the chosen data set does not cover enough chemical space for the downstream data sets, as seen in the model trained on ZINC in MolFormer \cite{ross2022large}. A systematic analysis of the impact of pre-training data set selection can be made by ensuring similarity to desired downstream data sets or by ensuring diversity within the pre-training data set itself \cite{albalak2024survey}.

\subsection{Which chemical language to use?}

% SMILES vs SELFIES
The transformer model is a sequence-based model that relies on learning relations within or between sequences. While molecules are 3D objects, they can be represented as graphs with atoms as nodes and bonds as edges. This molecular graph can be transcribed to produce a 1D representation, i.e., a sequence of atoms and bonds between them. 
The most common sequence-based small molecule languages are SMILES \cite{weininger1988smiles} and SELFIES \cite{krenn2020self}. 
The majority of the existing models have been trained exclusively using the SMILES representation (Table \ref{tab:pre-training_datasets}), except for ChemBERTa \cite{chithrananda2020chemberta}, RT \cite{born2023regression}, SELFormer \cite{yuksel2023selformer}, and MAT \cite{maziarka2020molecule}. 

The authors of ChemBERTa studied the impact of the choice of chemical language, i.e., SMILES vs. SELFIES, on performance when predicting binding to the stress response target p53 (SR-p53), a task from the Tox21 data set. They concluded that the difference was not significant (without showing further details). 
For RT, given that the model was designed as a decoder that relies on generation, the authors opted for the SELFIES representation as it is designed to be used in DL generation tasks to produce valid molecules in terms of syntax and properties \cite{krenn2020self, krenn2022selfies}. However, the authors made an initial comparison by training a base model with either SMILES or SELFIES and evaluated both of them for drug-likeness prediction. The results showed very close performance when reporting the RMSE metric. The models trained with SELFIES performed $\sim 0.004 \pm 0.01$ better on average. However, reporting the Pearson correlation coefficient metric reverses the conclusion with the SMILES model performing $\sim 0.004 \pm 0.01$ better on average.
Furthermore, the authors of SELFormer hypothesized that the validity of the SELFIES language makes it more suitable for representation learning than SMILES. 
The authors compared their model - trained on SELFIES - to the reported values of MolBERT \cite{fabian2020molecular} and ChemBERTa-2 \cite{ahmad2022chemberta} - trained on SMILES. 
SELFormer performed, e.g., better on the BBBP data set. However, as stated earlier, the need for more standardization of the data splitting makes comparisons inconclusive. 

\begin{figure}
  \includegraphics[width=1.0\linewidth]{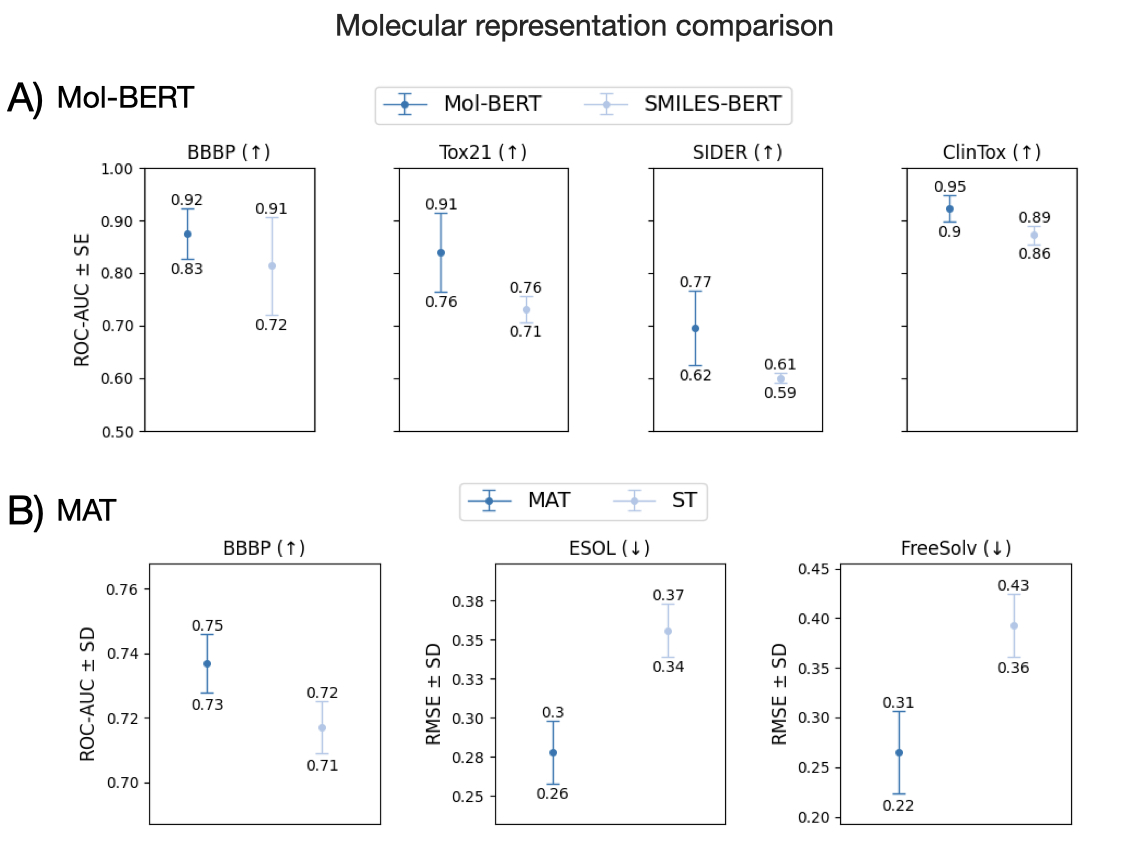}
  \caption{Comparison between the performance of the string-based models and models that used different representation inputs. \textbf{A)} A comparison between the Mol-BERT \cite{li2021mol} model trained on Morgan fingerprints of radius one and SMILES-BERT \cite{wang2019smiles} trained on SMILES. \textbf{B)} A comparison between the MAT \cite{maziarka2020molecule} model trained on a list of atoms and SMILES-Transformer (ST) \cite{honda2019smiles} trained on SMILES. The figure shows the average performance for each model with error bars (SE) or standard deviation (SD). The data and code used to generate this figure can be found in \href{https://github.com/volkamerlab/Transformers4MPP_review/tree/main}{our GitHub repo}.}
  \label{fig:representation}
\end{figure}

% Custom, domain-specific input (fingerprint-based sequence encodings, add pairwise 3d atom-distances, etc.)
Besides the SMILES vs. SELFIES question, Mol-BERT \cite{li2021mol} and FP-BERT \cite{wen2022fingerprints} opted for a different input representation. Instead of feeding the molecule to the model as an alphanumeric sequence, the authors exploited circular fingerprints like Morgan fingerprints and ECFP4 to be the input to the model. Both papers used a radius of one to convert the molecular graph into individual atoms and their first-degree neighborhood. The extracted substructures were then hashed and sorted by the atoms' order in the graph to form a sequence input for the model. FP-BERT \cite{wen2022fingerprints} compared their model to the reported values from MolBERT \cite{fabian2020molecular}. However, as stated earlier, the lack of standardization of the data splitting makes cross-comparisons inconclusive.
The Mol-BERT \cite{li2021mol} model performed a comparison analysis to the SMILES-BERT \cite{wang2019smiles} model on the BBBP, Tox21, SIDER, and ClinTox data sets. Figure \ref{fig:representation} A) shows that Mol-BERT outperforms SMILES-BERT on the SIDER data set with no overlapping error bars ($\sim 0.60 \pm 0.01$ and $\sim 0.67 \pm 0.07$, respectively). It also performs better than SMILES-BERT with minimal overlap in the error bar for Tox21 and ClinTox.

MAT \cite{maziarka2020molecule} is a model that modified the representation of a molecule by using only the list of its atoms without further grammatical rules as presented in SMILES or SELFIES. However, the authors integrate information about the inter-atomic 3D distances and adjacency in the molecular graph during self-attention calculation. The authors performed a comparison analysis with the ST \cite{honda2019smiles} model. Figure \ref{fig:representation} (B) shows that MAT performed better than ST on the three compared data sets, $\sim 0.737 \pm 0.009$  and $\sim 0.717 \pm 0.008$ ROC-AUC, respectively for BBBP, $\sim 0.278 \pm 0.020$ and $\sim 0.356 \pm 0.017$ RMSE, respectively for ESOL, and $\sim 0.265 \pm 0.042$ and $\sim 0.393 \pm 0.032$ RMSE, respectively for FreeSolv.

In conclusion, most of the models have used SMILES for training.
Only a few direct comparisons were performed when using SMILES vs SELFIES, for which no significant difference in performance was observed. The choice of the chemical language might depend on the application scenario of the model. For example, RT \cite{born2023regression} opted for SELFIES as the latter should be more suitable for generative models due to its ensured validity \cite{krenn2020self}. However, both ChemFormer \cite{irwin2022chemformer} and X-Mol \cite{xue2022x} adopted a decoder subunit while training on the SMILES language, and both reported almost 100\% valid molecules for molecular optimization tasks. These results suggest that the transformer model can encode meaningful information from either SMILES or SELFIES. Additionally, using a representation input that ignores syntax, such as in MAT \cite{maziarka2020molecule}, or a different representation as in Mol-BERT \cite{li2021mol}, was performing competitively to the SMILES-based models.

\subsection{How to tokenize the molecules?}

% tokenization in NLP
The transformer model learns an embedding for each token in an input sequence. Different tokenization methods can give a different granularity when representing the composing units of a language. The most popular tokenization schemes in NLP are character-level, subword-level, and word-level \cite{mielke2021between}. 
BERT \cite{devlin2018bert} and RoBERTa \cite{liu2019roberta} adopt similar techniques of subword tokenization where input is firstly tokenized by character. Then, the most common pairs of tokens keep getting merged until reaching a pre-defined vocabulary size. By doing so, a newly observed word during evaluation that is out of vocabulary (OOV) can be tokenized and embedded in terms of its composing subwords. 

\begin{table}[]
\begin{tabular}{lllll}
\hline
\textbf{Model} & \textbf{Tokenizer} & \textbf{Max Vocab} & \textbf{Max Sequence} & \textbf{Positional} \\
\textbf{Name} &  & \textbf{Size} & \textbf{Length} & \textbf{Encoding} \\
\hline
SMILES-BERT \cite{wang2019smiles} & Atom & NA & NA & Absolute \\
ST \cite{honda2019smiles} & Atom & None($^1$) & 220($^1$) & Absolute ($^2$) \\
MAT \cite{maziarka2020molecule} & Atom & NA & NA & Absolute ($^2$) \\
Chen et al. \cite{chen2021extracting} & Atom & 61 ($^3$) & 256 & Absolute ($^2$) \\
Transformer-CNN \cite{karpov2020transformer} & Atom & 66 & 110 & Absolute ($^1$) \\
MolBERT \cite{fabian2020molecular} & Atom & 42 & 128 & Relative \\
K-BERT \cite{wu2022knowledge} & Regex & 47 & 201 & Absolute ($^2$) \\
X-Mol \cite{xue2022x} & Regex & 112($^1$) & 512($^1$) & Absolute ($^2$) \\
ChemFormer \cite{irwin2022chemformer} & Regex & 523 & 512($^1$) & Absolute ($^2$) \\
MolFormer \cite{ross2022large} & Regex & 2,362 & 202 & Rotary \\
FP-BERT \cite{wen2022fingerprints} & Substructures & 3357 & 256 & Absolute ($^2$) \\
Mol-BERT \cite{li2021mol} & Substructures & 13,325 & 200($^1$) & Absolute ($^1$) \\
RT \cite{born2023regression} & Internal Tokenizer & 509 & 250 & Relative \\
SELFormer \cite{yuksel2023selformer} & BPE & 800($^1$) & 128($^1$) & Absolute ($^2$) \\
ChemBERTa \cite{chithrananda2020chemberta} & BPE, regex & 52K & 512 & Absolute ($^2$) \\
ChemBERTa-2 \cite{ahmad2022chemberta} & Common characters & 591 & NA& Absolute ($^2$) \\
\hline
\end{tabular}
\caption{An overview of tokenization methods and positional encoding by the different models. The "Tokenizer" column specifies how a sequence was split into tokens. The "Max Vocab Column" shows how many unique tokens were selected for the training. The "Max Sequence Length column" specifies the number of tokens in a sequence to be given as input to the model. The "Positional Encoding" column shows how information about the tokens' positions was added to the model. NA values mean that the authors did not mention it explicitly, did not share the code, or did not provide the code to replicate the results. Footnotes: $^1$ This info was not mentioned explicitly in the paper and was derived from the provided code. More information is found in \href{https://github.com/volkamerlab/Transformers4MPP_review/tree/main}{our GitHub repo}. $^2$ This information was not mentioned explicitly and is assumed to be the same as the base model mentioned in Table \ref{tab:articles}. $^3$ The authors mention 51 in the manuscript, but the supplementary table they refer to shows and counts 61 symbols}
\label{tab:tokenization_position}
\end{table}

% tokenization in MPP
Table \ref{tab:tokenization_position} shows which tokenizer was used for each reviewed model and the maximum number of tokens in the used vocabulary. Different tokenizers have been utilized for the SMILES language, such as the atom-level tokenizer, where the sequence is split into its composing atoms, numbers, and special characters. The regular expression (regex) tokenization was proposed by Schwaller et al. \cite{schwaller2018found}, which also segments the molecule into individual atoms. However, the characters within a square bracket and the double digits preceded by \% are treated as single characters. ChemBERTa \cite{chithrananda2020chemberta} and SELFormer \cite{yuksel2023selformer} utilized the byte-pair encoder (BPE) used in RoBERTa \cite{liu2019roberta}, which performs subword tokenization. Moreover, ChemBERTa \cite{chithrananda2020chemberta} attempted a comparison between the BPE and the regex tokenizers on the SR-p53 task of the Tox21 data set, where an improvement of 0.015 PRC-AUC for the regex tokenizer was reported. The RT \cite{born2023regression} model used the SELFIES internal tokenizer, which splits the molecule based on the language-specified symbols \cite{krenn2020self}. The only models that used a different tokenization method were Mol-BERT \cite{li2021mol} and FP-BERT \cite{wen2022fingerprints}, where the sequence was converted into substructures using circular fingerprints. 

% vocabulary size in MPP
The vocabulary size is a trainable parameter that adds to the model's complexity. Models like BERT \cite{devlin2018bert} and RoBERTa \cite{liu2019roberta} use vocabularies of sizes around 30K and 50K, respectively. It can be seen that, for the chemical language models, the vocabulary size varied dramatically from as little as 42 tokens for MolBERT \cite{fabian2020molecular} to over 52K tokens for ChemBERTa \cite{chithrananda2020chemberta} (Table \ref{tab:tokenization_position}). None of the articles provided further analysis of the effect of varying the vocabulary size on performance.

In conclusion, there is currently no detailed analysis available on which tokenization method may be most beneficial for which research question. However, it should be relevant to experiment with meaningful multi-character tokenizations as this provides the composing units of the sequence that the model will learn to represent \cite{mielke2021between}. A meaningful tokenization method can also allow us to judge the model's explainability. One possible approach for tokenization is to extend it to proper and informative fragmentation of the molecules by using methods relying on chemical rules (e.g., molBLOCKS \cite{ghersi2014molblocks}, eMolFrag \cite{liu2017break}, etc.) or data-driven methods (e.g., SMILES-base encoder (SPE) \cite{li2021smiles}, Morfessor \cite{creutz2005unsupervised}, etc.).

\subsection{How to add positional embeddings?}

% positional embeddings in NLP
The transformer \cite{vaswani2017attention} models rely on the attention mechanism to calculate the relationships between the tokens. However, this process is known to be position-agnostic \cite{yun2019transformers}. Therefore, information about the position of each token needs to be included so that the model can draw position-aware relationships. The original transformer \cite{vaswani2017attention} model provides an absolute and fixed positional encoding using different scaled sinusoidal functions. For each position in the sequence, a specific vector of the same length as the token's embedding is added to the token's embedding. Other models like GPT \cite{brown2020language}, BERT \cite{devlin2018bert}, and RoBERTa \cite{liu2019roberta} used a trainable absolute positional embedding so that the model would learn the embeddings of each position instead of having it pre-calculated. Fixed and trainable approaches are constrained by the sequence length when representing the absolute positions. Consequently, the model cannot assign a positional encoding if a longer sequence is encountered during evaluation. Therefore, other models \cite{dai2019transformer, su2021roformer} have attempted to learn a relative positional encoding, allowing for more flexible sequence processing during evaluation.

% positional embeddings in MPP 
Table \ref{tab:tokenization_position} shows the types of positional embedding used for the reviewed articles. All the models are inferred to use the same positional embedding as their base models, as detailed in Table \ref{tab:articles}. However, MolBERT \cite{fabian2020molecular}, RT \cite{born2023regression}, and MolFormer \cite{ross2022large} reported different methods for incorporating positional embedding. MolBERT \cite{fabian2020molecular} and RT \cite{born2023regression} adopted a relative positional embedding \cite{dai2019transformer} in which the position of a token is learned based on its position from the surrounding tokens. MolFormer \cite{ross2022large} opted for positional embeddings that combined both absolute and relative concepts, namely, Rotary Positional Embedding (RoPE) \cite{su2021roformer}. An advantage of the RoPE method is that it is compatible with the memory-efficient linear attention mechanism used in MolFormer \cite{ross2022large}. MolFormer \cite{ross2022large} performed a comparison between training their models with either absolute positional embedding or RoPE on the QM9 data set. Their preliminary results showed that the absolute embeddings performed better than the out-of-box implementation of RoPE. However, with their modified RoPE implementation, the authors found it to provide better performance when the pre-training data set size became huge. 
The authors report an average and standard deviation MAE over 13 regression tasks from the QM9 data set for models trained with 111K, 111M, and 1.1B molecules, respectively. The difference in performance between the absolute and RoPE positional embedding for these three models are $\sim -0.20 \pm 0.18$, $\sim -0.44 \pm 0.22$, and $\sim 0.27 \pm 0.12$, respectively. These results demonstrate that the RoPE embeddings were superior to absolute embeddings, but only when the pre-training data set was very large.    

% Custom, domain-specific positional embeddings or trying to add 2D/3D positonal info of atoms
An important aspect is that languages like SMILES and SELFIES are linearized versions of the 2D molecular graph.
Consequently, two consecutive tokens in sequence are not necessarily nearby in 2D or 3D representation. Therefore, encoding tokens' 2D or 3D positions might be more beneficial than encoding the positions in a SMILES sequence. The authors of MAT \cite{maziarka2020molecule} incorporated spatial information such as inter-atomic 3D distances and adjacency in the 2D molecular graphs. They did so by integrating such information alongside the self-attention mechanism. This approach performed comparably (Figure \ref{fig:representation} B) to the ST \cite{honda2019smiles} model, which uses absolute and fixed positional embeddings based on the SMILES sequence.

In conclusion, most of the papers have utilized the absolute positional embedding of their base model, which is either fixed or trainable, except for MolBERT \cite{fabian2020molecular}, RT \cite{born2023regression}, and MolFormer \cite{ross2022large}. These first analyses indicate that changing the absolute embedding helped when the pre-training data set was very large \cite{ross2022large}. Moreover, attention needs to be paid to how positional embeddings are used. Since molecules exist in 3D space, it can be challenging for linear chemical languages to fully represent the nature of molecules. For example, the SMILES representation is a linearization of the 2D molecular graph. Consequently, a positional embedding that accounts for the actual location of each token in 2D (the molecular graph) or 3D space would be an interesting direction to explore.

\subsection{How many parameters do we need?}

% model size in NLP
The transformer architecture consists of multiple parameters that need to be set and optimized for the underlying problem. For example, the transformed embeddings can be split between multiple attention heads inside an encoder or a decoder layer. By doing so, each head can attend separately to different dimensions of the tokens. Moreover, multiple transformer layers can be stacked to increase the depth of learning as each layer passes its learned representation of the tokens' embeddings to the next layer. Another parameter is the embedding length (usually referred to as the hidden size), as increasing it might correspond to higher expressiveness and generalization capabilities. The number of layers and the embedding length account for the total number of parameters to be learned by the model. Additionally, another set of parameters is encountered when learning the input embeddings to the model, which depends on the embedding length and the vocabulary size. The current trend in NLP has been towards increasing the number of parameters to allow for higher-quality learning by the model. For example, BERT \cite{devlin2018bert} was trained as a small model with 110M parameters and a large model with 230M parameters. The RoBERTa model was trained on 335M parameters, while GPT2 \cite{radford2019language} and GPT3 \cite{radford2019language} models were trained with 1.5B and 175B parameters, respectively. 

% model size in MPP
Table \ref{tab:parameters} shows that most of the chemical language models applied for MPP were trained with $\le$ 100M learnable parameters, which is in the small range of model sizes compared to models from NLP. The small size of the chemical language models might be understandable since languages like SMILES contain much smaller vocabulary and grammatical rules than natural languages. The table also shows how the choices of the individual hyperparameters varied between the different models. For example, ST \cite{honda2019smiles} embedded their tokens with 256 dimensions, while SMILES-BERT \cite{wang2019smiles} and MAT \cite{maziarka2020molecule} used 1024 dimensions for the embeddings. The number of layers varied from 4 to 32 by ST \cite{wang2019smiles} and MAT \cite{maziarka2020molecule}, respectively.

\begin{table}[]
\makebox[\textwidth]{\begin{tabular}{llllll}
\hline
\textbf{Model Name}   & \textbf{\# dimensions} & \textbf{\# head} & \textbf{\# layer} & \multicolumn{2}{c}{\textbf{\# parameters ($\sim$)($^1$)}} \\ \hline
 & & & & Model & Embeddings \\ \hline
ST \cite{honda2019smiles}                  & 256                  & 4               & 4                & 7M & NA   \\
Mol-BERT \cite{li2021mol}                  & 300                  & 6               & 6                & 7M & 4M \\
Transformer-CNN \cite{karpov2020transformer}& 512                 & 10              & 3                & 22M                     & 34K \\
Chen et al. \cite{chen2021extracting}      & 512                  & 8               & 8                                          & 25M                   & 26K \\
RT \cite{born2023regression}               & 256                  & 16              & 32               & 25M   &130K     \\
K-BERT \cite{wu2022knowledge}              & 768                  & 12              & 6                & 43M & 36K \\
ChemBERTa-2 \cite{ahmad2022chemberta}($^2$)& 768                  & 12              & 6                & 43M & 454K \\
ChemBERTa \cite{chithrananda2020chemberta} & 768                  & 12              & 6                & 43M & 40M  \\
ChemFormer \cite{irwin2022chemformer}      & 512                  & 8              & 6                & 44M  & 268K      \\
SMILES-BERT \cite{wang2019smiles}          & 1024                 & 4               & 6                & 76M & NA  \\
MolBERT \cite{fabian2020molecular}         & 768                  & 12              & 12               & 85M  & 32K 
\\
X-Mol \cite{xue2022x}                      & 768                  & 12              & 12               & 85M & 86K  \\
MolFormer \cite{ross2022large}             & 768                  & 12              & 6, 12            & 43M, 85M & 2M, 2M   \\
SELFormer \cite{yuksel2023selformer}       & 768                  & 12, 4           & 8, 12            & 57M, 85M  & 614K, 614K 
\\
MAT \cite{maziarka2020molecule}            & 1024                 & 16              & 8                & 101M  & NA \\
FP-BERT \cite{wen2022fingerprints}         & 256                  &   NA              &  NA                &  NA & NA           \\ 
\hline
\end{tabular}}
\caption{Some of the architecture choices for training the transformer blocks. 
The "\# dimensions" table corresponds to the embedding dimensions (also known as hidden size). The "\# head" corresponds to the number of heads specified per each attention block. The "\# layers" specifies the number of stacked encoders/decoders. The "\# parameters" corresponds to the number of learnable parameters by the models.
NA values mean that the authors did not mention it explicitly and either did not share the code or did not provide the code to replicate the results. 
Footnotes: $^1$ These values were calculated based on the provided information. The model's parameters depend on the model dimensions (hidden size) and the number of layers, while the embedding parameters depend on the vocabulary and hidden sizes. A script that does and explains the calculations can be found in  \href{https://github.com/volkamerlab/Transformers4MPP_review/tree/main}{our GitHub repo}. $^2$ ChemBERTa-2 \cite{ahmad2022chemberta} did not mention the corresponding parameters explicitly and were assumed to have the same parameters as ChemBERTa \cite{chithrananda2020chemberta}.}
\label{tab:parameters}
\end{table}

% Explicit evaluation of model size in the reviewed articles
As shown in Table \ref{tab:parameters}, only MolFormer \cite{ross2022large} and SELFormer \cite{yuksel2023selformer} provided MPP analysis for different sizes of their models. SELFormer \cite{yuksel2023selformer} varied the number of heads and the number of layers jointly and reported the performance on different classification and regression data sets. When fine-tuning both models using 50 epochs, the large model with $\sim 86M$ parameters performed better for some data sets (e.g., $\sim 0.034$ $\triangle ROC-AUC$ for BBBP) and very comparably for others (e.g., $\sim 0.001$ $\triangle ROC-AUC$ for BACE)   
A similar trend was observed for regression as the larger model performed $\sim 0.013$ $\triangle RMSE$ better for FreeSolv, while the improvement was only $\sim 0.007$ for PDBind.
However, the smaller model could perform $\sim 0.013$ better for LipoPhilicity. The previous observation shows that a larger model can improve the performance compared to a smaller model, which was also observed for most of the data sets for the MolFormer \cite{ross2022large} model. The larger model with $\sim 87M$ parameters performed $\sim 0.04$ ROC-AUC better on average than the smaller model ($\sim 57M$ parameters) for BBBP, HIV, BACE, and SIDER data sets. The larger model also performed better on FreeSolv ($\sim 0.12$ $\triangle RMSE$) and very comparably on ESOL ($\sim 0.001$ $\triangle RMSE$). The exception to the previous trend was noticed in the ClinTox and Tox21 data sets, as the larger model prominently outperformed the smaller model with as much as 0.34 and 0.42 ROC-AUC, respectively. 
  
In conclusion, the current state-of-the-art does not allow for systematical analysis of parameter scaling in transformer models for MPP. Preliminary analysis shows that larger models can provide some improvement. However, publications from the NLP domain show that scaling the model's parameters (aka, model size) should be carried out alongside the scaling of the training data set size \cite{kaplan2020scaling, hoffmann2022training}. Additionally, Kaplan et al. \cite{kaplan2020scaling} concluded that such scaling is more sensitive to the model size irrespective of the model shape (e.g., different models with the same number of parameters but different embedding lengths and the number of layers should behave similarly). However, Tay et al. \cite{tay2021scale} argue that model shape and size contribute to efficient downstream transfer. Therefore, proper systematic analysis of model scaling, similar to NLP, is still needed for the chemical language models to identify the most efficient choices. 

\subsection{Which pre-training objectives to adopt?}  

The transformer \cite{vaswani2017attention} models rely on self-supervised learning. Models developed with this scheme learn from labels that can be easily created from unlabeled data. Table \ref{tab:objectives} shows the objectives used for each reviewed article, which can be grouped into domain-agnostic and domain-specific. Most models used the same objectives as their base transformer model, originally built for NLP. Therefore, the objective is independent of the application area. However, models like MAT \cite{maziarka2020molecule}, MolBERT \cite{fabian2020molecular}, K-BERT \cite{wu2022knowledge}, and ChemBERTa-2 \cite{ahmad2022chemberta} used objectives that are specific to the chemical languages and general chemical information, hence, introducing inductive bias. 

\begin{figure}
  \includegraphics[width=1.0\linewidth]{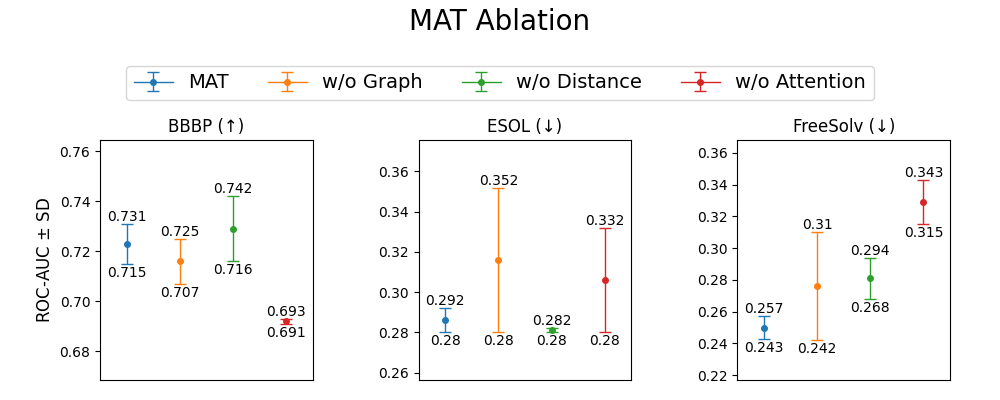}
  \caption{The ablation analysis by the MAT model\cite{maziarka2020molecule}. MAT = the model trained with the three sources of information, self-attention, interatomic 3D distances, and adjacency matrix. w/o graph = the MAT model trained with self-attention and interatomic distances only. w/o Distance = the MAT model with self-attention and adjacency only. w/o attention = the MAT model with interatomic distances and adjacency only. The data and code used to generate this figure can be found in  \href{https://github.com/volkamerlab/Transformers4MPP_review/tree/main}{our GitHub repo}. }
  \label{fig:mat_ablation}
\end{figure}

MAT \cite{maziarka2020molecule} used the interatomic 3D distances and the adjacency matrix of the molecular graph to integrate them as additional sources of information alongside the self-attention mechanism. In the ablation analysis, the authors removed one source of information at a time. They reported performance on BBBP, ESOL, and FreeSolv data sets as shown in Figure \ref{fig:mat_ablation}. The data sets behaved differently for each model. For example, the most stable performance for ESOL in terms of standard deviation was the model without the interatomic distance ( $\sim 0.28 \pm 0.001$ RMSE). However, the model trained with the three objectives was the most stable across the three data sets. 

MolBERT \cite{fabian2020molecular} introduced two new objectives alongside MLM: PhysChemPred and SMILES-EQ. The former objective entails training the model to predict $\sim$ 200 physico-chemical properties. At the same time, the latter classifies a pair of SMILES as equivalent or not, given the non-uniqueness feature of the SMILES language. In their analysis of virtual screening outcomes, the authors found that the PhysChemPred objective performed very closely to the model trained with the three objectives, $\sim 0.72 \pm 0.06$ and $\sim 0.71 \pm 0.06$ ROC-AUC, respectively (Supplementary Figure \ref{fig:molber_ablation}). 
ChemBERTa-2 \cite{ahmad2022chemberta} incorporated the same PhysChemPred objective, naming it Multiple Task Regression (MTR), and compared its performance to a model trained with MLM only. Figure \ref{fig:size_performance} (B) shows that the model trained with MTR-only performed noticeably better for almost all four classification tasks and pre-training data set size combinations than MLM-only.
The SMILES-EQ objective proposed by MolBERT \cite{fabian2020molecular} was found to lower the performance when combined with the other objectives, $\sim 0.71 \pm 0.06$ and $\sim 0.74 \pm 0.07$ ROC-AUC for the models with and without the SMILES-EQ objective, respectively.
However, K-BERT \cite{wu2022knowledge} incorporated the non-uniqueness of the SMILES language as well but as a contrastive learning (CL) objective. With this objective, the model is trained to maximize the similarity between the SMILES sequences of the same molecule while minimizing similarities to the other sequences. In their ablation analysis, the model trained with and without CL performed similarly for the 15 tested data sets (Supplementary Figure \ref{fig:k-bert_ablation}) with an average ROC-AUC of $0.806$ and $0.807$, respectively.
The authors used two other objectives: predicting features of heavy atoms (e.g., hybridization) and the molecule's MACCS keys descriptor. However, no further ablation analysis was performed for these two objectives. 

In conclusion, most models utilized the domain-agnostic pre-training objectives as their base transformer model. The remaining models that used inductive bias showed it can boost performance. This line of research might be interesting since chemical languages like SMILES are not self-contained, but other factors influence molecular properties. Thus, further physico-chemical properties or 3D conformations might be needed to improve the prediction performance of individual properties. Additionally, it would be interesting to quantify the extent of what the language model captures independently to understand better which objectives would further enhance the model's performance and explainability. For example, visualizing the attention scores calculated for the tokens during training and fine-tuning can help to understand what the model is paying attention to and whether it is relevant to the downstream data set \cite{rogers2021primer}.  Payne et al. \cite{payne2020bert} performed such an analysis by training a BERT-based model to quantify the model's ability to capture functional groups. They visualized the attention scores and leveraged chemical knowledge to judge whether the model paid proper attention to well-described functional groups.

\begin{table}[]
\begin{tabular}{lll}
\hline
\textbf{Model Name}                        & \textbf{Pre-training objective ($^1$)} & \textbf{Fine-tuning}                 \\ \hline
X-Mol \cite{xue2022x}                      & -    & Update\\
ST \cite{honda2019smiles}                  & -    & Frozen\\
FP-BERT \cite{wen2022fingerprints}         & MLM  & Frozen\\
MolFormer \cite{ross2022large}             & MLM  & Frozen, Update\\
SELFormer \cite{yuksel2023selformer}       & MLM  & Frozen, Update\\ 
SMILES-BERT \cite{wang2019smiles}          & MLM  & Update\\
ChemBERTa \cite{chithrananda2020chemberta} & MLM  & Update\\
Mol-BERT \cite{li2021mol}                  & MLM  & Update \\
Chen et al. \cite{chen2021extracting}      & MLM (Guided) & Update\\
RT($^2$) \cite{born2023regression}         & PLM  & - \\
Transformer-CNN \cite{karpov2020transformer} & Augmentation & Update \\
ChemFormer \cite{irwin2022chemformer}      & Denoising Span MLM, Augmentation     & Update \\
MAT \cite{maziarka2020molecule}            & MLM, 2D Adjacency, 3D Distance & Update\\
MolBERT \cite{fabian2020molecular}         & MLM, PhysChemPred, SMILES-EQ & Frozen, Update\\
ChemBERTa-2 \cite{ahmad2022chemberta}      & MLM, MTR                 & Update                \\
K-BERT \cite{wu2022knowledge}              & Atom feature prediction, MACCS prediction, CL & Update last layer \\
\hline
\end{tabular}
\caption{A list of the pre-training objectives and the fine-tuning choices made by the reviewed articles. 
Footnotes: $^1$ MLM = Masked Language Modelling. Guided MLM refers to masking chemically relevant tokens (e.g., functional groups). PLM = Permutation Language Modelling. PhysChemPred = Physico-chemical properties prediction for a molecule. SMILES-EQ = SMILES equivalence, identifying whether two SMILES represent the same molecule. CL = Contrastive learning, the task of learning representations by contrasting positive and negative samples. MTR = Multiple Task Regression (same as PhysChemPred). $^2$ Regression Transformer was trained with the aim of performing multi-task learning during pre-training. Therefore, they did not perform fine-tuning.}
\label{tab:objectives}
\end{table}

\subsection{How to fine-tune?}

% fine-tuning with frozen weights.
After optimizing the model's architecture and training on large unlabelled data sets, it can be used for a wide range of downstream tasks. Therefore, a pre-trained model can be a generic basis for multiple applications. The output representation of the model can be used directly as a proper numeric descriptor, a so-called embedding, of the input \cite{peters2019tune}. When the downstream data set is closely related to the pre-training data, the output representation can be used as the input for a classifier/regressor model for the said data set. In this scenario, the model's weights are frozen and do not go through any further updating phase \cite{peters2019tune, min2023recent}. 
% fine-tuning with weight-updates 
The other method of fine-tuning is to update the weights by appending a classification/regression neural network head to predict the output of the underlying downstream data set \cite{peters2019tune, min2023recent}. In this scenario, one or more of the model's layers get updated to maximize the performance gain on the corresponding labeled data set. Updating the weights requires further hyperparameter tuning to find the best learning convergence \cite{min2023recent}. The full model update has been used in many publications \cite{vaswani2017attention, devlin2018bert}. However, this causes further computational toll. Some methods from NLP have been proposed to enable efficient fine-tuning by updating specific relevant weights or biases or even updating a small initialized new weight matrix and adding it to the frozen original model \cite{min2023recent}.   

% explicitly tried fine-tuning in MPP
Table \ref{tab:objectives} shows the different fine-tuning methods used by the discussed models. ST \cite{li2021smiles} and FP-BERT \cite{wen2022fingerprints} were the only two models that used the frozen fine-tuning strategy alone, while the rest used either the update strategy only or compared the performance of both. We assume the full model updating was performed for the updating strategy since no contradicting information is provided in the manuscripts. Only K-BERT \cite{wu2022knowledge} stated that they updated the weights of the last layer of the model. MolBert \cite{fabian2020molecular} and SELFormer \cite{yuksel2023selformer} compared their models using the two strategies on multiple downstream data sets, while MolFormer \cite{ross2022large} tested the two strategies on the QM9 data set only. 
Tabel \ref{tab:finetune} shows the difference in performance between frozen and updated models of MolBERT \cite{fabian2020molecular} and SELFormer \cite{yuksel2023selformer} for three regression data sets. The table shows that updated models performed better than the frozen model (e.g., a difference in the performance of 0.575 and 2.187 $\triangle RMSE$ for MolBERT and SELFormer, respectively, for the FreeSolv data set). However, the SELFormer \cite{yuksel2023selformer} model seems to benefit from weight updates much more than MolBERT \cite{fabian2020molecular} for the three shown data sets in table \ref{tab:finetune}. This observation might be explained by the pre-training objectives used in MolBERT that might have improved the learned representations before fine-tuning. For example, Payne et al. \cite{payne2020bert} show that a certain layer in the fine-tuned model on the solubility data set has learned to attend to the long carbon chain in the molecule. This observation relates well with the LogP (partition coefficient), which indicates the relationship between lipophilicity and solubility. The pre-training objectives in MolBERT \cite{fabian2020molecular} included physico-chemical properties, such as LogP, making the pre-trained model closer to some downstream data sets.
Another observation from Table \ref{tab:finetune} is that SELFormer Lite, which is a smaller version of SELFormer in terms of the model's parameters, is seen to benefit from weight updates fine-tuning more than the larger model for ESOL (1.311 and 0.871 $\triangle RMSE$, respectively). 

In conclusion, fine-tuning can be done in multiple ways \cite{min2023recent}, where the current literature investigated a couple of approaches. The analysis shows that fine-tuning through weight updates provides better performance than frozen weights, which is understandable since fine-tuning by weight updates further directs the model to output embeddings that are more related to the downstream task. Additionally, domain-specific pre-training objectives and architectural choices, such as the number of parameters, can also influence the performance of the fine-tuned model.

\begin{table}[]
\begin{tabular}{llll} \hline
                 & \multicolumn{3}{c}{\textbf{$\triangle $ (Frozen - Updated) RMSE}}      \\ \hline
                 & \textbf{ESOL} & \textbf{FreeSolv} & \textbf{Lipophilicity} \\ \hline
MolBERT \cite{fabian2020molecular}          & 0.021       & 0.575           & 0.041   \\
SELFormer \cite{yuksel2023selformer}        & 0.971       & 2.187           & 0.310   \\
SELFormer Lite \cite{yuksel2023selformer}   & 1.311       & 2.112           & 0.388   \\ \hline
\end{tabular}
\caption{The advantage of fine-tuning through weight updates vs frozen weights. SELFormer and SELFormer Lite differ in the number of parameters. Random split was used for the three datasets by both models. The weight update strategy provides better results. However, some models seem to benefit more from this strategy than others. The results were reported for regression only as SELFormer compared fine-tuning strategies using random split while MolBERT used scaffold split.}
\label{tab:finetune}
\end{table}

\section{Current challenges for MPP with transformers}

% summary of what we tried to do:
In this review, we collected and assessed the performance of the currently ubiquitous transformer models for molecular property prediction. By categorizing the approaches and ideas from the literature, we found varying handling of benchmark data sets and reporting transparency to be the main obstacles. 

% the big problem of comparability
We observed that different test sets are often used by different publications (see Section \ref{sota}), which makes proper comparison of different models infeasible due to the lack of standardization. Firstly, not all models adhered to the recommended scaffold split for the data sets, which is supposed to test the model's generalizability better than a random split. Secondly, many papers used different splitting modules with different random seeds and sometimes various split ratios. Thus, each model was tested on a different test set, precluding proper comparison. Therefore, it is absolutely crucial to have a unified test split for the recommended splitting method so that benchmarking new and old models becomes accessible and more meaningful.

Given the broad collection of labeled property data sets (see Section \ref{downstream data}), selecting the most relevant ones is difficult. For example, the commonly used data sets from MoleculeNet or TDC and their experimental property data are heterogeneous. The data set sizes and compositions differ and might represent the respective endpoints in varying quality and generality. For example, contained molecules might cover only a limited chemical space, the property distribution is narrow or does not cover the practically relevant range. In addition, other relevant molecular properties are not covered, e.g., because of a lack of data or expensive manual curation for raw data processing. This can make benchmark collections not sufficiently represent real-world scenarios and inadequately describe the practical usefulness of prediction models. We believe that a high-quality description of the downstream tasks would help to focus and foster method development in the future.      

% outlook for representation evaluation
In this direction, some studies propose more detailed and systematic comparisons of molecular representations \cite{zhang2022can, deng2023systematic}. They focus on the well-known concepts of molecular scaffolds and "activity cliffs" and describe frameworks to analyze and identify desired representations for MPP. A good representation should be able to generalize across structurally different molecules (called "generalized scaffold hopping" \cite{zhang2022can} and "inter-scaffold generalization" \cite{deng2023systematic}) and be able to accurately represent small structural molecule modifications which result in considerable property changes (referred to as "generalized activity cliffs" \cite{zhang2022can} and "intra-scaffold generalization" \cite{deng2023systematic}). There are already benchmark platforms dedicated to activity cliffs, like MoleculeACE \cite{van2022exposing}.

% what should be reported about transformer methods in publications:
We enlisted a set of questions in this review that we believe are important for designing an efficient transformer model for MPP (see Section \ref{sec-decisions}). However, extracting the relevant information from each article was not always easy. Moreover, fishing for the needed information from the provided code by some articles was also very challenging. We believe the list of questions we provide in this review is a good start for agreeing on what should be reported in future works to enable meaningful comparisons and meta-analysis. We also believe that providing a structured and well-documented code repository for the community would help navigate the code and extract further useful information. 

% Which metric(s) to report 
We almost found a consensus on model evaluation metrics between the articles, as most used ROC-AUC for classification and RMSE or MAE for regression (see Section \ref{sota}). These metrics were also proposed in the MoleculeNet \cite{wu2018moleculenet} paper as the authors also observed the community using ROC, PRC, RMSE, and MAE. They further vouched for the selection of ROC or PRC based on the severity of class imbalance for classification data sets. For example, PRC-AUC is considered a more suited alternative for strongly imbalanced data sets \cite{saito2015precision, wu2018moleculenet}. Although a single metric is easy to compare and draw insights upon, it can hardly capture the complete picture \cite{hand2009measuring}. Therefore, it might be of interest for the community to consider multiple metrics to provide complementary insights about the performance \cite{robinson2020validating}. 

% statistical analysis
Further, metric values alone are usually insufficient to comprehensively judge the model's performance (see Section \ref{sota}). The next step should be performing statistical analysis to emphasize further that a performance improvement is consistent and robust \cite{nicholls2014confidence}. Statistical analysis requires multiple and independent runs of the evaluation step and the selection of the most appropriate measure to calculate significance \cite{nicholls2014confidence, nicholls2016confidence}. As only a few reviewed articles performed statistical analysis, we believe the community needs proper statistical analysis \cite{lehmann1986testing} to provide robust and meaningful conclusions.   

% Report visualization (a.k.a. no bold table)
After a proper evaluation, visualizing performance is crucial to engage the reader and allow other researchers to form an insightful understanding. Most reviewed papers report their findings as a table or bar plot. A table format requires the reader to mentally visualize the performance of the different models, which is challenging. Additionally, a bar plot cannot visualize continuous data as different distributions can correspond to the same bar plot \cite{weissgerber2017data}. For multiple evaluation runs, a summary plot like a boxplot will be more engaging and informative than a scatter plot \cite{weissgerber2017data, midway2020principles}. Moreover, plotting each data point is recommended when the number of runs is relatively small, as the summary drawn from a small number of data points might not be representative \cite{weissgerber2017data}.   

\section{Conclusion and outlook}

This work reviews the articles that trained a sequence-based transformer model for molecular property prediction (MPP). Our analysis showed that the transformer models have not unleashed unprecedented performance as seen in other fields like NLP \cite{vaswani2017attention} (Figure \ref{fig:performance}). However, they have shown promising performance compared to the existing MPP models.  
In an attempt to identify the factors that might aid in designing better transformer models for MPP, we explored various questions to which we found the following preliminary answers:

\begin{itemize}
    \item There are large enough data sets of unlabeled molecules that can be used for pre-training, e.g. ChEMBL, PubChem, ZINC, and even public and proprietary fragment spaces ranging up to $10^{26}$ molecules \cite{warr2022exploration}. However, results indicate that size alone is not the sole determinant of successful downstream predictions, but the data set composition, i.e., the covered chemical space, is equally relevant. Therefore, it is important to determine the relevant factors of pre-training data set composition that help to train predictive and generalizable models efficiently. Future directions might include the clustering and filtering of large databases while maintaining prediction performance on downstream tasks.
    \item Chemical languages such as SMILES and SELFIES are one way of linearizing the molecular graph while introducing some grammatical rules that add semantic information. Preliminary analysis shows that both languages are similarly sufficient for training a transformer model. Additionally, the input does not need to be strictly limited to these representations as other approaches can also work well when the model's architecture is adjusted accordingly \cite{wen2022fingerprints, li2021mol, maziarka2020molecule}. 
    \item The tokenization of the input sequence is an important step that determines the granularity of the model training. However, different approaches have not yet been systematically compared, and most groups have used character-based tokenization techniques. Future works can investigate the benefit of chemically and biologically meaningful tokenization for the downstream performance and explainability benefit.  
    \item The positional encoding has not undergone intensive analysis. The atoms of a molecule in the molecular graph have different relative positions in the generated SMILES or SELFIES. Therefore, questioning the proper way to incorporate positional embedding that reflects a token's 2D or 3D position is worth further studies. 
    \item In the NLP field, it has already been studied that model size (number of parameters) should be scaled concurrently with the data size \cite{kaplan2020scaling, hoffmann2022training}. 
    %The current literature on chemical language models does not show evidence of systematic model size scaling. 
    In some of the reviewed articles, scaling the model parameters was performed, but the data set size was fixed. Therefore, systematic analyses, like those in NLP, are needed to understand how the chemical language models behave under these constraints. 
    \item Pre-training objectives like masked language modeling (MLM) are mostly domain-agnostic. Multiple papers explored the effect of incorporating domain-specific objectives and comparisons showed a promising trend. However, some domain-specific objectives can be learned from the domain-agnostic objectives \cite{payne2020bert}. Therefore, further analysis is desirable to identify the needed pre-training objectives. 
    \item Fine-tuning a pre-trained model by weight updates is needed when the downstream data differs considerably from the pre-training data, which can be computationally expensive. Some discussed studies compared the frozen strategy to weight updates and reported the superior performance of the latter. However, we observe that architectural and pre-training choices can affect the added value of this step.
    Future work is needed to compare different fine-tuning strategies with different models to identify when the expensive step of weight updates is needed and how to perform it most efficiently.
    \item The literature's current performance evaluation process lacks important factors that enable proper comparisons. The community must ensure, document, and provide appropriate splitting for the downstream data set and a corresponding fixed test set for each split. Furthermore, statistical and significance analysis is highly needed, which is a vital factor for faithfully judging a model's benefit. 
    \item Visualization reporting is also one way of engaging the reader and allowing further valuable insights. We believe that reporting needs to move from a simple table to a proper visualization of the different runs and their distribution summary (e.g., boxplots).
    \item This review focused on architectural and domain-relevant exploration for training a transformer model. Addressing further aspects like weight initialization and selection of hyperparameters for both pre-training and fine-tuning steps would also be helpful for future work.
    \item Additionally, one of the main advantages of the transformer models is their ability to capture long-range dependencies, which is usually identified through attention visualization of the trained and fine-tuned models \cite{rogers2021primer}. Therefore, reviewing the explainability analysis done by the different models can help to identify and highlight the added benefit of the transformer model. It can also help to shed light on the weaknesses that must be addressed to improve the model's performance further.
    \item The current literature on chemical language models often focuses on approaches and ideas from natural language processing. The NLP field is moving fast and there are multiple new ideas and applications \cite{kaddour2023challenges} that may improve chemical model performance, including pre-training regimes \cite{wang2022text, nussbaum2024nomic}, fine-tuning strategies \cite{gunel2020supervised, hu2021lora}, explainability  \cite{zhao2024explainability} and more. At the same time, the requirements for chemical language models differ from those in NLP. For example, chemical models have smaller vocabularies and lower sequence lengths (see Table \ref{tab:tokenization_position}), and not all concepts need to be transferred from text to molecules. Therefore, adapting models to the specificities of chemical data necessary to accurately predict the properties of molecules needs to be explored in the future. 
\end{itemize}

In conclusion, the transformer models show promising performance for molecular property prediction. We believe that they can be further improved when accounting for the identified best practices and recommendations mentioned above. Finally, we believe the transformer's context-dependent embeddings can improve the intricate association of the molecule's structure with molecular properties.
In addition, the self-supervised learning scheme can potentially improve small labeled tasks. Both characteristics make transformers a promising addition to existing models.
Therefore, we believe that further improvements in the transformer model to match the requirements of molecular property prediction will provide better performance, generalizability, and explainability.  

\begin{acknowledgement}
The authors thank Soheila Samiee and Jesujoba Oluwadara Alabi for valuable discussions that improved the manuscript. 
\end{acknowledgement}

\bibliography{bibliography}

\appendix

\begin{table}[]
\makebox[\textwidth]{\begin{tabular}{llllllll}
\hline
Model Name                                 & \textbf{BBBP}       & \textbf{HIV} & \textbf{BACE} & \textbf{Tox21} & \textbf{ClinTox} & \textbf{SIDER} & \textbf{Others} \\ \hline
MolFormer \cite{ross2022large}             & \checkmark & \checkmark   & \checkmark    & \checkmark     & \checkmark       & \checkmark     & - \\
ST \cite{honda2019smiles}                  & \checkmark          & \checkmark   & \checkmark    & \checkmark     & \checkmark       & -              & - \\
SELFormer \cite{yuksel2023selformer}       & \checkmark & \checkmark   & \checkmark    & \checkmark     & -                & \checkmark     &  -  \\
Transformer-CNN \cite{karpov2020transformer}& \checkmark          & \checkmark            & \checkmark             & \checkmark              & \checkmark   & -              & \begin{tabular}[c]{@{}l@{}}AMES, JAK3,\\BioDeg, RP AR.\end{tabular} \\ 
Mol-BERT \cite{li2021mol}                  & \checkmark & -   & -    & \checkmark              & \checkmark                 & \checkmark              & -  \\
X-Mol \cite{xue2022x}                    & \checkmark & \checkmark   & \checkmark    & -              & \checkmark       & -              & MUV. \\
ChemBERTa \cite{chithrananda2020chemberta} & \checkmark          & \checkmark   & -             & \checkmark ($^1$)    & \checkmark ($^1$)     & -              & - \\
MolBERT \cite{fabian2020molecular}         & \checkmark & \checkmark   & \checkmark    & -              & -                 & -              & - \\
Chen et al. \cite{chen2021extracting}& \checkmark          & -            & \checkmark             & -              & \checkmark   & -              & AMES, BEET.   \\  
ChemBERTa-2 \cite{ahmad2022chemberta}      & \checkmark & -             & -             & \checkmark ($^1$)    & -\checkmark ($^1$)      & -              & -  \\
FP-BERT \cite{wen2022fingerprints}         & \checkmark & \checkmark   & -             & -              & -                & -              &  - \\
MAT \cite{maziarka2020molecule}            & \checkmark          & -            & -             & -              & - & -              & \begin{tabular}[c]{@{}l@{}}ESTROGEN-a, \\ ESTROGEN-ß, \\ METSTABLow, \\ METSTABHIGH. \end{tabular} \\
SMILES-BERT \cite{wang2019smiles}          & -                   & -            & -             & -              & -                & -              & \begin{tabular}[c]{@{}l@{}}LogP, PM2, \\ PCBA 686978. \end{tabular} \\             ChemFormer \cite{irwin2022chemformer}      & -           & -            & -             & -              & -  \\
K-BERT \cite{wu2022knowledge}              & -          & -            & -             & -               & -                & -               & \begin{tabular}[c]{@{}l@{}}Pgb-sub, HIA, \\ F(20\%), F(30\%), \\ FDAMDD, \\ CYP1A2-sub, \\ CYP2C19-sub, \\ CYP2C9-sub, \\ CYP2D6-sub, \\ CYP3A4-sub, \\ T12, DILI, \\ SkinSen, \\ Carcinogenicity, \\ Respiratory, \\ Malaria. \end{tabular} \\
RT \cite{born2023regression} ($^2$)          & -          & -            &  -            & -              & -                & -             & -  \\ \hline
\end{tabular}}
\caption{The classification data sets used by some of the models. These models used scaffold split except for X-Mol, Trnasformer-CNN, and Chen et al. 
Footnotes: $^1$ a single task was predicted for this data set instead of all tasks as done by the other models. $^2$ Regression Transformer (RT) was trained for predicting regression data sets only.} 
\label{tab:data_calssification}
\end{table}

\begin{table}[]
\begin{tabular}{lllll}
\hline
\textbf{Model Name}                        & \textbf{ESOL} & \textbf{FreeSolv} & \textbf{Lipophilicity} & \textbf{Others}          \\ \hline

ST \cite{honda2019smiles}                  & \checkmark    & \checkmark        & \checkmark     & -                \\
MolBERT \cite{fabian2020molecular}         & \checkmark    & \checkmark        & \checkmark     & -                \\
Chen et al. \cite{chen2021extracting}     & \checkmark    & \checkmark        & \checkmark     & LogS, DPP4.      \\
ChemFormer \cite{irwin2022chemformer}      & \checkmark    & \checkmark        & \checkmark     & -                \\
X-Mol \cite{xue2022x}                      & \checkmark    & \checkmark        & \checkmark     & -                \\
FP-BERT \cite{wen2022fingerprints}         & \checkmark    & \checkmark        & \checkmark     & Malaria, CEP.    \\
MolFormer \cite{ross2022large}             & \checkmark    & \checkmark        & \checkmark     & QM9, QM8.        \\
RT \cite{born2023regression}               & \checkmark    & \checkmark        & \checkmark     & -                \\
SELFormer \cite{yuksel2023selformer}       & \checkmark    & \checkmark        & \checkmark     & PDBind.          \\
MAT \cite{maziarka2020molecule}            & \checkmark    & \checkmark        & -              & -                \\
Transformer-CNN \cite{karpov2020transformer}& -            & \checkmark        & \checkmark     & \begin{tabular}[c]{@{}l@{}}MP, BP, BCF, LogS, \\BACE, DHFR, LEL. \end{tabular}               \\
SMILES-BERT \cite{wang2019smiles}          & -             & -                 & -              & -                \\
ChemBERTa \cite{chithrananda2020chemberta} & -             & -                 & -              &  -               \\
Mol-BERT \cite{li2021mol}                  & -             & -                 & -              & -                \\
K-BERT \cite{wu2022knowledge}              & -             &  -                & -              & -                \\
ChemBERTa-2 \cite{ahmad2022chemberta}      & -             & -                 & -              & Clearence, Delaney, BACE. \\
\hline
\end{tabular}
\caption{The regression data sets modeled by some of the models. These models used random split.}
\label{tab:data_regression}
\end{table}

\begin{table}[]
\begin{tabular}{lll} \hline
End Point    & \% Positive & \# Molecules \\ \hline
CT\_TOX            & 8.0         & 1484        \\
FDA\_APPROVED      & 94.0        & 1484        \\ \hline
\end{tabular}
\caption{The percentage of positive class per task for the ClinTox data set. FDA\_approved corresponds to compounds that have been approved by the FDA. CT\_TOX corresponds to compounds that failed in clinical trials due to toxicity reasons. Data and code used to generate this table can be found in \href{https://github.com/volkamerlab/Transformers4MPP_review/tree/main}{our GitHub repo}.}
\label{tab:clintox_pos_pct}
\end{table}

\begin{table}[]
\makebox[\textwidth]{\begin{tabular}{lll} \hline
End Point                                                           & \% Positive & \# Molecules \\ \hline
Product issues                                                      & 2.0         & 1427        \\
Pregnancy, puerperium and perinatal conditions                      & 9.0         & 1427        \\
Surgical and medical procedures                                     & 15.0        & 1427        \\
Congenital, familial and genetic disorders                          & 18.0        & 1427        \\
Social circumstances                                                & 18.0        & 1427        \\
Endocrine disorders                                                 & 23.0        & 1427        \\
Neoplasms benign, malignant and unspecified (incl cysts and polyps) & 26.0        & 1427        \\
Ear and labyrinth disorders                                         & 46.0        & 1427        \\
Reproductive system and breast disorders                            & 51.0        & 1427        \\
Hepatobiliary disorders                                             & 52.0        & 1427        \\
Eye disorders                                                       & 61.0        & 1427        \\
Blood and lymphatic system disorders                                & 62.0        & 1427        \\
Renal and urinary disorders                                         & 64.0        & 1427        \\
Injury, poisoning and procedural complications                      & 66.0        & 1427        \\
Cardiac disorders                                                   & 69.0        & 1427        \\
Metabolism and nutrition disorders                                  & 70.0        & 1427        \\
Infections and infestations                                         & 70.0        & 1427        \\
Musculoskeletal and connective tissue disorders                     & 70.0        & 1427        \\
Psychiatric disorders                                               & 71.0        & 1427        \\
Immune system disorders                                             & 72.0        & 1427        \\
Respiratory, thoracic and mediastinal disorders                     & 74.0        & 1427        \\
Vascular disorders                                                  & 78.0        & 1427        \\
Investigations                                                      & 81.0        & 1427        \\
Nervous system disorders                                            & 91.0        & 1427        \\
General disorders and administration site conditions                & 91.0        & 1427        \\
Gastrointestinal disorders                                          & 91.0        & 1427        \\
Skin and subcutaneous tissue disorders                              & 92.0        & 1427        \\ \hline
\end{tabular}}
\caption{The percentage of positive class per task for the SIDER data set. Each row corresponds to compounds that have been affiliated with a certain side effect. Data and code used to generate this table can be found in \href{https://github.com/volkamerlab/Transformers4MPP_review/tree/main}{our GitHub repo}.}
\label{tab:sider_pos_pct}
\end{table}

\begin{table}[]
\begin{tabular}{lll} \hline
End Point     & \% Positive & \# Molecules \\ \hline
NR-PPAR-gamma & 3.0         & 6450        \\
NR-AR         & 4.0         & 7265        \\
NR-AR-LBD     & 4.0         & 6758        \\
SR-ATAD5      & 4.0         & 7072        \\
NR-Aromatase  & 5.0         & 5821        \\
NR-ER-LBD     & 5.0         & 6955        \\
SR-HSE        & 6.0         & 6467        \\
SR-p53        & 6.0         & 6774        \\
NR-AhR        & 12.0        & 6549        \\
NR-ER         & 13.0        & 6193        \\
SR-ARE        & 16.0        & 5832        \\
SR-MMP        & 16.0        & 5810        \\ \hline
\end{tabular}
\caption{The percentage of positive class per task for the tox21 data set. Each row corresponds to compounds that interfere with a certain cellular pathway.}
\label{tab:tox21_pos_pct}
\end{table}

\begin{table}[]
\makebox[\textwidth]{\begin{tabular}{lllll}
\hline
\textbf{Model Name}                        & \multicolumn{2}{c}{\textbf{Metric}} & \textbf{Visualization}  & \textbf{Statistics} \\ \hline
 & \textbf{Classification}  & \textbf{Regression} &  &   \\ \hline
MAT \cite{maziarka2020molecule}	           & ROC	                                                  & RMSE	  & \begin{tabular}[c]{@{}l@{}}Average $\pm$ sd,\\Rank boxplot.\end{tabular} 	               & -  \\
MolBERT \cite{fabian2020molecular}         & ROC                                                     & RMSE     & Average $\pm$ sd                                 & -  \\
ChemBERTa-2 \cite{ahmad2022chemberta}      & ROC                                                    & RMSE      & Average                                          & -  \\
SELFormer \cite{yuksel2023selformer}       & ROC                                                    & RMSE      & Average                                          & -  \\ 
X-Mol \cite{xue2022x}                      & ROC                                                    & RMSE     
 & Bar Plot.                                       & One-tailed t-test \\
FP-BERT \cite{wen2022fingerprints}         & ROC/PRC      & RMSE, R$^2$, Q$^2$ & Average $\pm$ sd  & Paired t-test($^1$)\\
Transformer-CNN \cite{karpov2020transformer}             & ROC                                                    & R$^2$  & Bar Plot $\pm$ se                               & -  \\
Chen et al. \cite{chen2021extracting}             & ROC                                                    & R$^2$  & Bar Plot $\pm$ se                               & -  \\
MolFormer \cite{ross2022large}             & ROC                                                    & RMSE, MAE($^1$)  & Average                               & -  \\
ST \cite{honda2019smiles}                  & \begin{tabular}[c]{@{}l@{}}ROC/PRC,\\DEM\end{tabular}  & RMSE, DEM   & \begin{tabular}[c]{@{}l@{}}Average,\\Scatter plot $\pm$ sd \end{tabular}  & -  \\
ChemBERTa \cite{chithrananda2020chemberta} & ROC, PRC                                                & -          & Average                                       & -  \\
Mol-BERT \cite{li2021mol}                  & ROC, F1                                                & -          & Average $\pm$ sd                              & -  \\
K-BERT \cite{wu2022knowledge}              & ROC, MCC($^1$)                                         & -           & Average $\pm$ sd                             & -  \\
SMILES-BERT \cite{wang2019smiles}          & ACC                                                         & -         & Bar Plot                                    & -  \\
ChemFormer \cite{irwin2022chemformer}      & -                                                           & RMSE, R$^2$   & Average                              & -   \\
RT \cite{born2023regression}               & -                                                           & RMSE, MAE($^1$)  & Average $\pm$ sd                  & -   \\
\hline                      
\end{tabular}}
\caption{The evaluation choices made by the reviewed articles. sd = standard deviation and se = standard error. ACC = Accuracy, DEM = Data efficient metric which is proposed by ST to quantify the model's robustness against different evaluation data set sizes, ROC = Receiver operating characteristic, PRC = Precision-Recall curve, MCC = Mathew's correlation coefficient, RMSE = Root-Mean-Square Error, MAE = Mean Absolute Error. $^1$ Some metrics were used for only a subset of the analysis.}
\label{tab:evaluation}
\end{table}

\begin{figure}
  \includegraphics[width=1.0\linewidth]{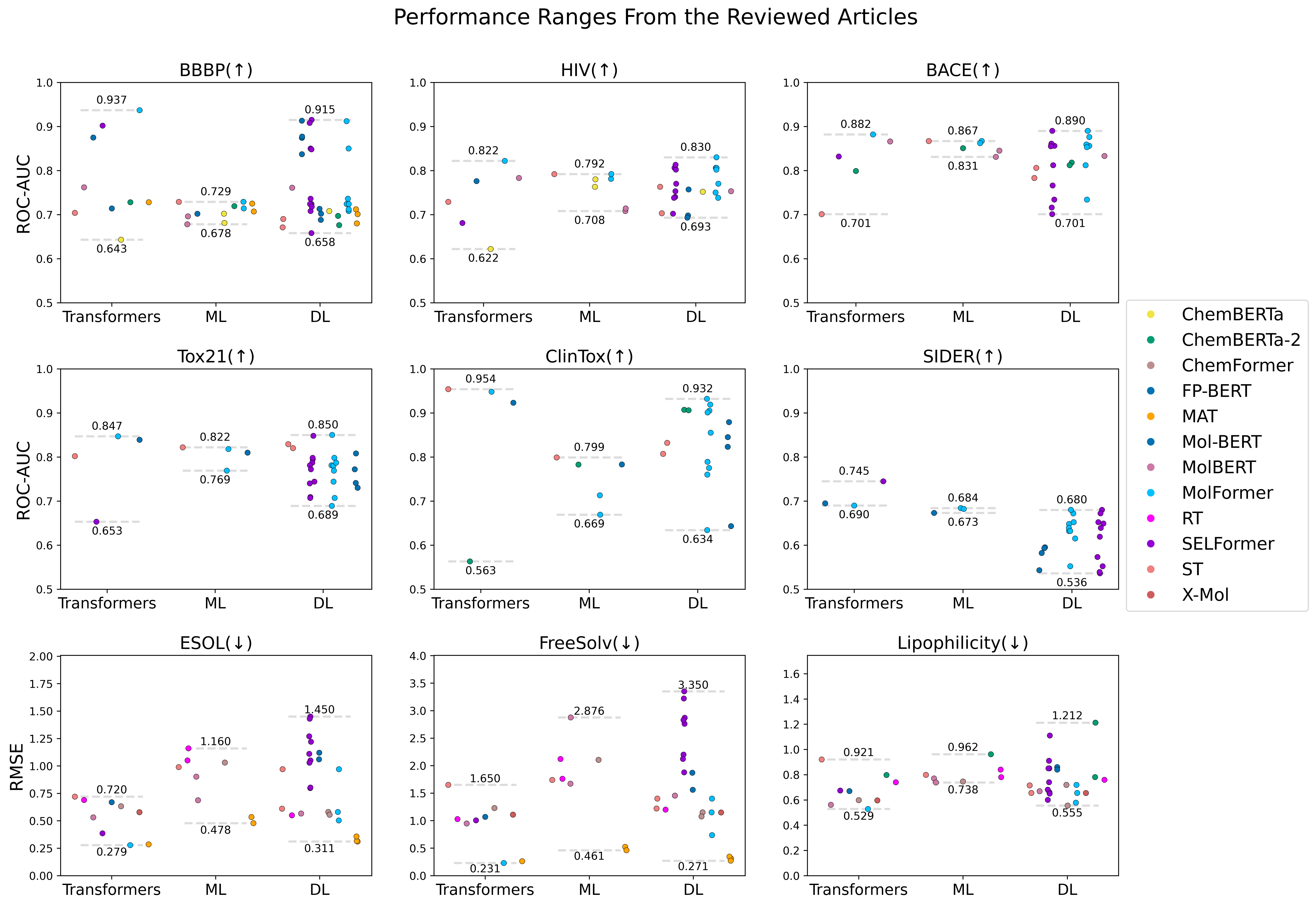}
  \caption{A comparison between the ROC-AUC and RMSE ranges for the reviewed articles, some classical machine learning (ML) algorithms, and some deep learning (DL) models. Scaffold splitting was used for the classification data sets and random splitting was used for the regression data sets by all the models. However, the test set is not guaranteed to be unified across the different models and categories. 
  The values range for the transformers models span the corresponding models in Table \ref{tab:data_regression}. The reported classical ML category spans models like RF, SVM, etc. The DL models span graph-based and different DNN models like D-MPNN, Weave, etc. Supplementary Table \ref{tab:models_comparisons} shows which classical ML and DL models were used for comparison by each transformer model. Data and code used to generate this figure can be found in  \href{https://github.com/volkamerlab/Transformers4MPP_review/tree/main}{our GitHub repo}.}
  \label{fig:performance}
\end{figure}

\begin{table}[]
\begin{tabular}{lll}
\hline
            & Classical ML                    & DL                \\ \hline
MAT \cite{maziarka2020molecule}         & ECFP + (SVM, RF)                & GCN, Weave, EAGCN \\
MolBERT \cite{fabian2020molecular}     & RDKit descriptors or ECFP + SVM & CDDD + SVM        \\
Mol-BERT \cite{li2021mol}    & -                               & MPNN              \\
ChemBERTa \cite{chithrananda2020chemberta}   & Morgan FP + (SVM, RF)           & D-MPNN            \\
ChemBERTa-2 \cite{ahmad2022chemberta} & NA + RF ($^1$)                  & D-MPNN, GCN     
\\ \hline
\end{tabular}
\caption{This table shows which models were compared against each transformer model in Figure \ref{fig:performance_honest}. Footnotes: $^1$ The models did not report the corresponding info (i.e., the input for the ML model).}
\label{tab:models_comparable_comparisons}
\end{table}

\begin{table}[]
\makebox[\textwidth]{\begin{tabular}{lll}
\hline
            & Classical ML                    & DL                                                                                \\ \hline
ST \cite{honda2019smiles}          & ECFP + MLP                      & GraphConv, Weave                                                                  \\
X-Mol($^1$) \cite{xue2022x}    & …                               & MPNN, GC, …                                                                       \\
ChemFormer \cite{irwin2022chemformer}  & Morgan FP + (SVM, RF)           & D-MPNN, MPNN                                                                      \\
MAT \cite{maziarka2020molecule}         & ECFP + SVM                      & GCN, Weave, EAGCN                                                                 \\
MolBERT \cite{fabian2020molecular} & RDKit descriptors or ECFP + SVM & CDDD + SVM                                                                        \\
Mol-BERT \cite{li2021mol}    & ECFP + NA($^2$)                        & GraphConv, Weave, MPNN, FP2VEC                                                    \\
FP-BERT \cite{wen2022fingerprints}    & -                               & FP2VEC, FCNN, ByPass                                                              \\
MolFormer \cite{ross2022large}   & NA + (RF, SVM)($^2$)                   & MGCN, D-MPNN, DimeNet, Hu et al, \\
 & & N-Gram, MolCLR, GraphMPV-C, \\
 & & GeomGCL,   GEM, A-FP \\
ChemBERTa \cite{chithrananda2020chemberta}   & Morgan FP + (SVM, RF)           & -                                                                                 \\
ChemBERTa-2 \cite{ahmad2022chemberta} & NA + RF($^2$)                          & D-MPNN, GCN                                                                       \\
SELFormer \cite{yuksel2023selformer}   & -                               & CLM, Hu et al, MolCLR, GraphMPV-C, \\
 & & GEM, MGCN, GCN, GIN, SchNet, KPGT              \\
RT \cite{born2023regression}         & RF, XGBoost                     & MPNN      \\ \hline                                                               
\end{tabular}}
\caption{This table shows which models were compared against each transformer model in Figure \ref{fig:performance}. 
Footnotes: $^1$ The X-Mol model compared itself to the reported results of multiple models. However, they reported performance as a bar plot and showed numeric values for the best two models only. Therefore, only models with available numeric values were plotted. $^2$ The models did not report the corresponding info for the question mark (i.e., the ML model or the input source).}
\label{tab:models_comparisons}
\end{table}

\begin{figure}
  \includegraphics[width=1.0\linewidth]{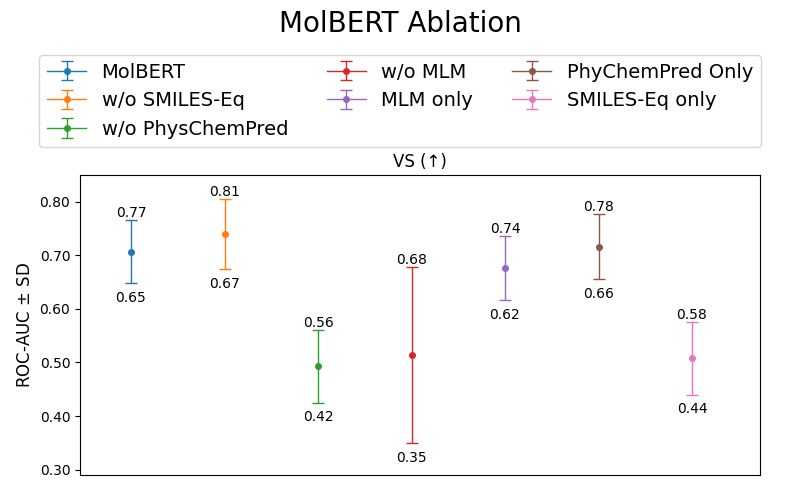}
  \caption{Visualization of the ablation analysis performed by MolBERT \cite{fabian2020molecular} on virtual screening data set. The authors implemented three training objectives and reported performance for all combinations of them. The plot shows the average performance and standard deviation of repeated testing. The data and code used to generate this figure can be found in  \href{https://github.com/volkamerlab/Transformers4MPP_review/tree/main}{our GitHub repo}.}
  \label{fig:molber_ablation}
\end{figure}

\begin{figure}
  \includegraphics[width=1.0\linewidth]{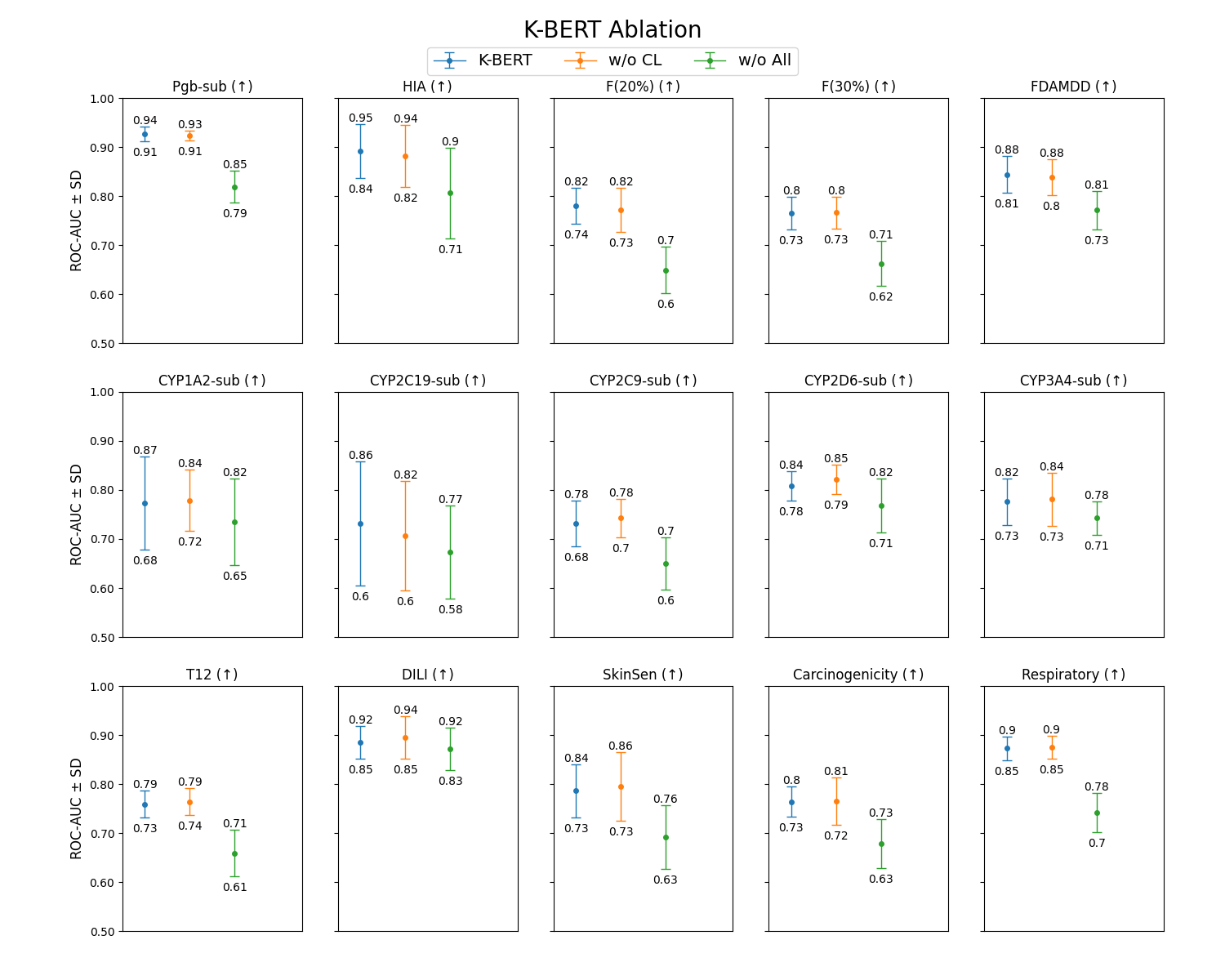}
  \caption{Visualization of the ablation analysis performed by K-BERT \cite{wu2022knowledge} on 15 pharmaceutical data sets. The authors implemented three training objectives, atom features prediction, MACCS keys prediction, and contrastive learning (CL). The authors reported ablation analysis only for the model trained without CL (w/o CL) and with no pre-training objectives (w/o All). The plot shows the average performance and standard deviation of repeated testing. The data and code used to generate this figure can be found in  \href{https://github.com/volkamerlab/Transformers4MPP_review/tree/main}{our GitHub repo}.}
  \label{fig:k-bert_ablation}
\end{figure}

\end{document}